\documentclass[10pt,journal,compsoc]{IEEEtran}

\ifCLASSOPTIONcompsoc
  \usepackage[nocompress]{cite}
\else
  \usepackage{cite}
\fi
\ifCLASSINFOpdf
  \usepackage[pdftex]{graphicx}
  \usepackage[caption=false,font=footnotesize]{subfig}
  \usepackage{multirow}
\else
\fi
\usepackage{amsmath}
\begin{document}
%
\title{Improving Shadow Suppression for Illumination Robust Face Recognition}
%
%
%
%

\author{Wuming~Zhang,
         Xi~Zhao,
        Jean-Marie Morvan
        and~Liming~Chen,~\IEEEmembership{Senior~Member,~IEEE}
\IEEEcompsocitemizethanks{\IEEEcompsocthanksitem W. Zhang and L. Chen are with the Laboratory of LIRIS, Department of Mathematic and Computer Science, Ecole Centrale de Lyon, Ecully, 69310 France.\protect\\ 
E-mail: \{wuming.zhang, liming.chen\}@ec-lyon.fr 
\IEEEcompsocthanksitem X. Zhao is with School of Management, Xi'an Jiaotong University, Xi'an 710049, China.\protect\\
E-mail: zhaoxi1@hotmail.com 
\IEEEcompsocthanksitem J. M. Morvan is with Universit\'{e} Lyon 1, Institut Camille Jordan, CNRS UMR 5208, 43 blvd du 11 Novembre 1918, F-69622 Villeurbanne-Cedex, France, and King Abdullah University of Science and Technology, Visual Computing Center, Bldg 1, Thuwal 23955-6900, Saudi Arabia.\protect\\
E-mail: jean-marie.morvan@kaust.edu.sa}}

\IEEEtitleabstractindextext{%
\begin{abstract}
2D face analysis techniques, such as face landmarking, face recognition and face verification, are reasonably dependent on illumination conditions which are usually uncontrolled and unpredictable in the real world. An illumination robust preprocessing method thus remains a significant challenge in reliable face analysis. In this paper we propose a novel approach for improving lighting normalization through building the underlying reflectance model which characterizes interactions between skin surface, lighting source and camera sensor, and elaborates the formation of face color appearance. Specifically, the proposed illumination processing pipeline enables the generation of Chromaticity Intrinsic Image (CII) in a log chromaticity space which is robust to illumination variations. Moreover, as an advantage over most prevailing methods, a photo-realistic color face image is subsequently reconstructed which eliminates a wide variety of shadows whilst retaining the color information and identity details. Experimental results under different scenarios and using various face databases show the effectiveness of the proposed approach to deal with lighting variations, including both soft and hard shadows, in face recognition.
\end{abstract}

\begin{IEEEkeywords}
Face recognition, lighting normalization, illumination and texture analysis
\end{IEEEkeywords}}

\maketitle

\IEEEdisplaynontitleabstractindextext

%
\IEEEpeerreviewmaketitle

\ifCLASSOPTIONcompsoc
\IEEEraisesectionheading{\section{Introduction}\label{sec:introduction}}
\else
\section{Introduction}
\label{sec:introduction}
\fi

%
%
%
%
\IEEEPARstart{F}{ace} analysis has received a great deal of attention due to the enormous developments in the field of biometric recognition and machine learning. Beyond its scientific interest, face analysis offers unmatched advantages for a wide variety of potential applications in commerce and law enforcement as compared to other biometrics, such as easy access or avoidance of explicit cooperation from users \cite{zhao2003face}. Nowadays conventional cases have attained quasi-perfect performance in a highly constrained environment wherein poses, illuminations, expressions and other non-identity factors are strictly controlled. However these approaches suffer from a very restricted range of application fields due to non-ideal imaging environments frequently encountered in practical cases: the users may present their faces not with a neutral expression, or human faces come with unexpected occlusions such as sunglasses, or even the images are captured from video surveillance which can gather all the difficulties such as low resolution images, pose changes, lighting condition variations, etc. In order to be adaptive to these challenges in practice, both academic and industrial research understandably shift their focus to unconstrained real-scene face images. 

\begin{figure}
\subfloat[]{
\label{fig:improved_subfig_a}
\centering
\includegraphics[width=0.225\linewidth]{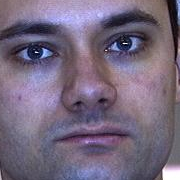}}
\subfloat[]{
\label{fig:improved_subfig_b}
\centering
\includegraphics[width=0.225\linewidth]{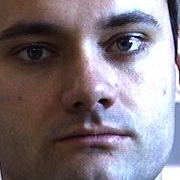}}
\subfloat[]{
\label{fig:improved_subfig_c}
\centering
\includegraphics[width=0.225\linewidth]{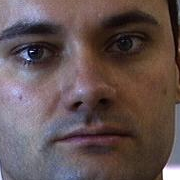}}
\subfloat[]{
\label{fig:improved_subfig_d}
\centering
\includegraphics[width=0.225\linewidth]{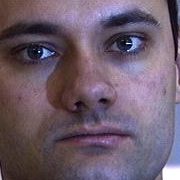}}
\caption{An example of varying lighting conditions for the same face. (a) Front lighting; (b) Specular highlight due to glaring light coming from right side; (c) Soft shadows and (d) hard-edged cast shadow.}
\label{lightings}
\end{figure}

Compared with other nuisance factors such as pose and expression, illumination variation impinges more upon many conventional face analysis algorithms which assume a normalized lighting condition. As depicted in Fig. \ref{lightings}, the lighting condition can be fairly complicated due to numerous issues: the intensity and direction of the lighting, the overexposure and underexposure of the camera sensor, just to name a few. Not only that, but it has already been proven that in face recognition, differences caused by lighting changes could be even more significant than differences between individuals \cite{adini1997face}. Therefore, lighting normalization turns out to be crucially important for exploring illuminant-invariant approaches. 

\begin{figure*}[t]
\centering
\includegraphics[width=\linewidth]{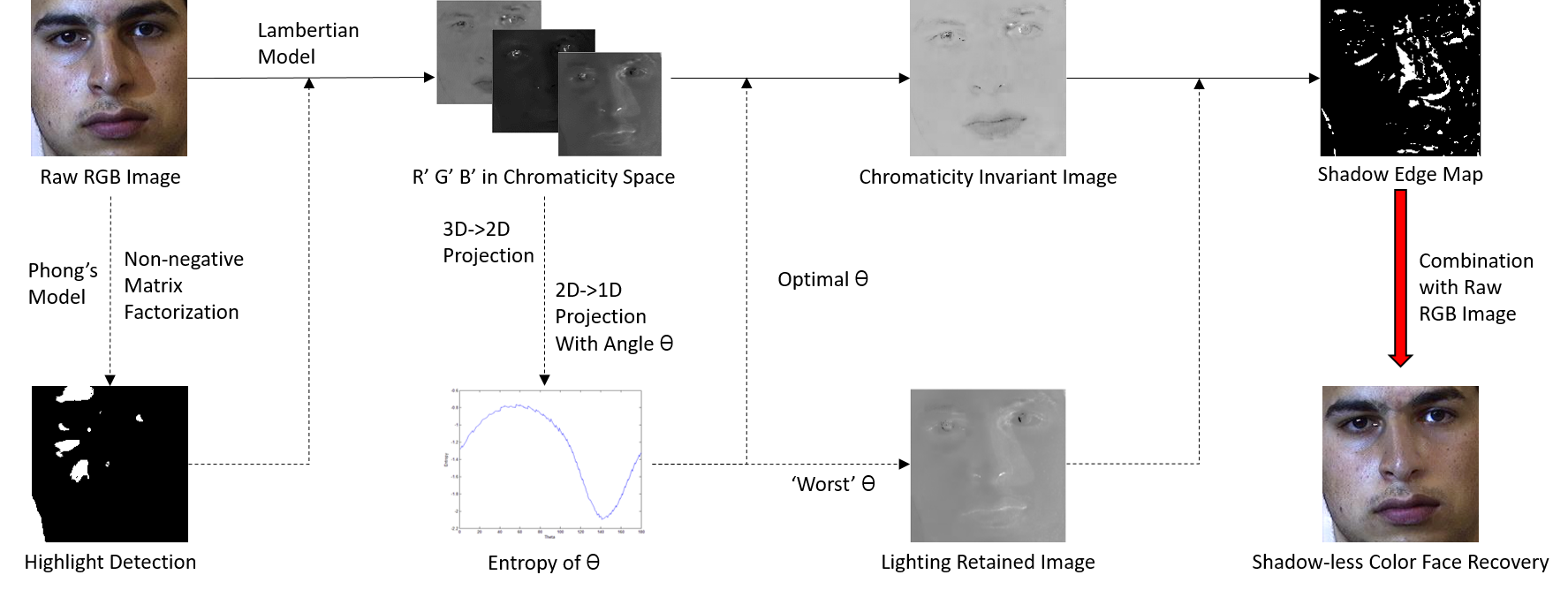}
\caption{Overview of the chromaticity space based lighting normalization process and shadow-free color face recovery process.}
\label{overview}
\end{figure*}


In this paper, we first divide the whole face into highlighted and non-highlighted regions; then the approximations of Lambertian surfaces and Planckian lighting are made to investigate the image formation rules; a pixel-level transformation in log space which aims at pursuing a chromaticity invariant representation is afterwards constructed; and the final step is to extend this chromaticity invariance to color space by taking into account the shadow edge detection. An overview of the proposed processing method is illustrated in Fig. \ref{overview}. Ultimately the experiments are carried out based on lighting normalized images and favorable experimental results have been achieved on the CMU-PIE and the FRGC face database. Our specific contributions are listed as follows.

\begin{enumerate}
\item We introduce and develop a chromaticity-based physical interpretation for modeling the face imaging process, which takes highlight detection as preprocessing and is able to separate the effect of illumination from intrinsic face reflectance.
\item We present a novel application of chromaticity invariant image for shadow-free color face reconstruction rather than gray-scale level de-lighting, demonstrating the potential to recover photo-realistic face image while eliminating the lighting effect.
\item We evaluate the proposed method on two benchmarking datasets across illumination variations and demonstrate that it can help improve performance of state-of-the-art methods especially on hard shadows, both qualitatively and quantitatively.
\end{enumerate}

The remainder of this paper is structured as follows: Section 2 briefly overviews related work in illumination invariant face recognition; Section 3 describes the color formation principles of human faces in RGB space while Section 4 details an illumination-normalized intrinsic image formation algorithm in chromaticity space; in Section 5 this invariance is further studied to enable full color shadow-free face recovery; promising experimental results and conclusions are given respectively in Section 6 and Section 7.


\section{Related Work}
Over the years, a surge of qualitative and quantitative studies on illumination invariant research have been put forward by reason of their suitability and efficacy in face analysis.  These techniques could be roughly divided into three categories according to their diverse theoretical backgrounds: holistic normalization methods, invariant feature extraction methods and 3D model based methods.

Holistic normalization based approaches used to be common in early algorithms. They attempt to redistribute the intensities of the original face image in a more normalized representation, which is less prone to lighting changes by applying a simple gray-scale intensity adjustment. Histogram Equalization (HE) and Histogram Matching (HM) \cite {pizer1987adaptive} initiated these methods by adopting an image preprocessing stage on the histogram level. Shan et al. \cite {shan2003illumination} developed Gamma Intensity Correction (GIC) for normalizing the overall image intensity at the given illumination level by introducing an intensity mapping:  $G(x,y)=cI(x,y)^{1/\gamma}$ where $c$ is a gray stretch parameter and $\gamma$ is the Gamma coefficient. Notwithstanding their ease of implementation and the apparent beneficial effects on lighting normalization, these methods fail to further satisfy the more and more rigorous demands on accuracy because they are global and do not take into account the in-depth image formation principles, which means that they only average the holistic intensity distribution and can hardly handle soft shadow, hard shadow or highlight respectively.

In view of this deficiency of holistic normalization, invariant feature extraction methods are conducted. The extraction of illumination-invariant components from the frequency domain is a mainstream idea which yields the implementation of wavelet-based denoising \cite{zhang2009multiscale} and logarithmic discrete cosine transform (LDCT) \cite{chen2006illumination}. Derived from Land's Retinex model \cite{land1971lightness} and its variants, which indicated that a face image could be decomposed into its smoothed version and its illumination invariant features, Riklin-Raviv and Shashua \cite {riklin1999quotient} proved that Quotient Image (QI), i.e. a ratio image between test image and a linear combination of three prototype images based on the Lambertian model, is illumination free. The algorithm is then generalized by Wang et al. \cite {wang2004generalized} in the Self Quotient Image (SQI) which replaced the prototype images by a smoothed version of test image itself. SQI achieved predominant performance while suffering from lack of edge-preserving capability caused by their weighted Gaussian filter. Chen et al. \cite {chen2006total} utilized the TV-$L^1$ model for factorizing an image and succeeded in overcoming this drawback. Local Normalization (LN) \cite{xie2006efficient} was proposed by Xie et al. to cope with uneven lighting conditions by reducing or even removing the effect of noise. Gradientface \cite{zhang2009face} and Weberface \cite{wang2011illumination} compute the ratio of $x$-gradient to $y$-gradient and the ratio of the local intensity variation to the background of a given image, respectively, to obtain illumination invariant representations. An integrative preprocessing chain was performed by Tan and Triggs \cite {tan2010enhanced} which successively merged Gamma correction, Difference of Gaussian filtering, optional masking and contrast equalization. All these approaches achieved impressive performance on removing soft shadows, yet encountered problems with hard-edged cast shadows especially caused by self-occlusion around the nose. Meanwhile, these techniques can not be extended to color space, resulting in limited applications in real world.

With the ever-advancing development of 3D data acquisition and application technologies, many researchers turned their attention to 3D model estimation for dealing with lighting problems based upon physical principles. Basri et al. \cite{basri2003lambertian} proved that a convex Lambertian object obtained under a large variety of lighting conditions can be approximated by a 9D linear subspace. Blanz and Vetter \cite {blanz2003face} first proposed the 3D Morphable Model (3DMM) to estimate and synthesize lighting conditions by means of linear combination of prototype models. A publicly available 3D Morphable Face Model - the Basel Face Model (BFM) \cite{paysan20093d} - was then constructed to realize the widespread use of 3DMM. Wang et al. \cite {wang2009face} presented Spherical Harmonic Basis Morphable Model (SHBMM) fusing 3DMM and spherical harmonic illumination representation \cite{basri2003lambertian}. Based on physical lighting models, Zhao et al. \cite {zhao2014minimizing} decomposed lighting effects using ambient, diffuse, and specular lighting maps and estimated the albedo for face images with drastic lighting conditions. 3D based lighting independent methods are powerful and accurate compared with 2D based ones. However they are easily confined to data acquisition and the unavoidable high computational cost. Even we can compromise by considering only 2D images and normalizing their lightings using 3D models, data registration between 2D and 3D remains likewise an inconvenience.

In summary, the proposed approach in this paper, which is actually a fusion of holistic normalization and reflectance model, introduces for the first time, the usage of the chromaticity invariant image into the field of face analysis to reconstruct an shadow-free color face image without using 3D priors. Compared with existing methods, we have constructed a comprehensive framework which combines the physical interpretation of face imaging and the simplicity of implementation. Moreover, since the proposed method removes shadow in color space, it can jointly work with other gray-scale level techniques to improve lighting normalization performance.

\section{Skin Color Analysis}
In this section, we formulate a physics-based reflectance model for approximating pixel based face skin colors. To begin with, we recapitulate the definition and properties of the two most commonly used reflectance models, then a non-negative matrix factorization (NMF) based method is implemented to locate the highlighted facial region which is less informative for precise model formulation. A product-form representation which could account for diffuse color is finally proposed as the cornerstone for our approach.

\subsection{Reflectance Model: Lambert vs. Phong}
Despite the appearance of several more comprehensive and more accurate BRDF models such as Oren-Nayar \cite{oren1994generalization} and Hanrahan-Krueger \cite{hanrahan1993reflection} in recent years, they are practically constrained by computational burden and become heavily ill-posed with respect to inverse estimation of material reflectance which greatly restricts their application in general lighting normalization tasks. Instead, classical models like Lambert and Phong \cite{phong1975illumination} still occupy a prime position in this field due to their ease of implementation.

As a common assumption, Lambert and Phong both adopt the concept of ideal matte surface obeying Lambert's cosine law where the incident lighting arriving at any point of object surface is uniformly diffused in all observation directions. Furthermore, Phong's model extends Lambertian reflectance by adding a specular highlight modelisation term which is merely dependent on the object's geometric information and lighting direction at each surface point. The representation of the Lambertian model and Phong's model could be formulated by equation \eqref{Lambert} and \eqref{Phong}, respectively,
\begin{equation}\label{Lambert}
L_{diffuse} = S_d E_d (\boldsymbol{n \cdot l})
\end{equation}
\begin{equation}\label{Phong}
L_{diffuse} + L_{specular} = S_d E_d (\boldsymbol{n \cdot l}) + S_s E_s (\boldsymbol{v \cdot r})^{\gamma}
\end{equation}
where $S_d$ and $S_s$ denote the diffuse and specular reflection coefficients; $E_d$ and $E_s$ represent the diffuse and specular lighting intensities; $\boldsymbol{n}$, $\boldsymbol{v} $, $\boldsymbol{l}$ and $\boldsymbol{r} = 2\boldsymbol{(n \cdot l)n-l}$ refer to the normal vector, the viewer direction, the direction of incident light and the direction of the perfectly reflected ray of light for each surface point; $\gamma$ is a shininess constant.

Despite the fact that the human face is neither pure Lambertian (as it does not account for specularities) nor entirely convex, the simplifying Lambertian assumption is still widely adopted in face recognition studies \cite{belhumeur1998set, basri2003lambertian, ramamoorthi2001relationship, wen2003face, zhang2005face} as the face skin is mostly a Lambertian surface \cite{kee2000illumination}. Nevertheless, premising the work on this assumption would be suboptimal because the specular highlight is widely occurring in practice and could not be ignored in face images due to the inevitable existence of the oil coating and semi-transparent particles in the skin surface. To address this problem, we decide to first detect the highlight region on each face image using Phong-type model; the classical Lambertian reflectance will then be applied afterwards to the skin color analysis for the non-highlighted region.

\subsection{Specular Highlight Detection}


As was proven in \cite{madooei2015detecting}, the variations in density and distribution of skin pigments, such as melanin and hemoglobin, simply scales the skin reflectance function, i.e. $S_d(\boldsymbol{x},\lambda) = \beta(\boldsymbol{x})S_d(\lambda)$. Here $\boldsymbol{x}$ denotes the spatial coordinates. Furthermore, as stated in \cite{stan2005handbook}, spectrum of surface-reflected light for specular spots in face skin can be considered to be equal to the spectrum of source lighting, i.e. $S_s=1$, otherwise $S_s=0$ for non-highlighted regions. With these caveats in mind, each component in Phong's model could be divided into an achromatic term (decided only by geometric parameters) and a chromatic term (parametrized by $\lambda$):
\begin{equation}\label{Phong_DRM}
L(\boldsymbol{x},\lambda) = (\boldsymbol{n \cdot l})\beta(\boldsymbol{x})E_d(\lambda)+(\boldsymbol{v \cdot h})^{\gamma}S_s(\boldsymbol{x})E_s(\lambda)
\end{equation}

More specifically, the RGB responses could be rewritten as spatial coordinates determined by geometrical dependency in space spanned by the color of light and the color of surface:
\begin{equation}\label{RGB}
\begin{bmatrix}
R(\boldsymbol{x}) \\
G(\boldsymbol{x}) \\
B(\boldsymbol{x}) \\
\end{bmatrix}
= 
\begin{bmatrix}
R_d & R_s \\
G_d & G_s \\
B_d & B_s \\
\end{bmatrix}
\times
\begin{bmatrix}
k_d(\boldsymbol{x}) \\
k_s(\boldsymbol{x})
\end{bmatrix}
\end{equation}
where the first term of the right-hand side is a 3$\times$2 matrix representing RGB channel magnitudes for diffuse and specular reflection while the second achromatic term is a 2$\times$N matrix (N denotes the number of pixels) containing diffuse and specular coefficients.

\begin{figure}[t]
\includegraphics[width=\linewidth]{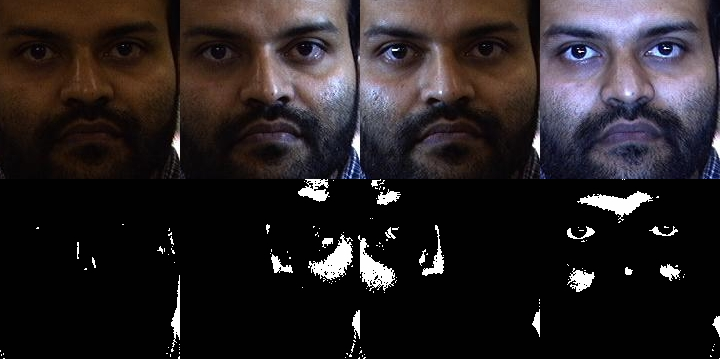}
\caption{Specular highlight detection results on images under various lighting conditions. Top row: original images; bottom row: detected highlight masks.}
\label{highlight}
\end{figure}

Remarkably, all these matrices are non-negative and $k_s(\boldsymbol{x})$ is sparse due to the fact that only a small portion of face contains specularity. It then becomes natural to consider the use of Non-negative Matrix Factorization (NMF) \cite{hoyer2004non} for solving such a $\boldsymbol{V = W \cdot H}$ problem. The implementation is easy: we set the inner dimension of factorization to 2 and apply a sparse constraint for $k_s(\boldsymbol{x})$ by restricting its $L_1$ norm while fixing its $L_2$ norm to unity as a matter of convenience. 
z
As demonstrated in Fig. \ref{highlight}, the performance of highlight detection using the proposed method for face images under different illumination environments is proved to be robust irrespective of lighting intensity and lighting direction.

\subsection{Skin Color Formation}
After successfully separating the surface-reflected region from body-reflected region, our focus will be to investigate the skin color formation on the dominant non-highlighted area using Lambertian reflectance model. Conceptually, there exist three primary factors which may be involved in a comprehensive image formation scene: source lighting, object surface and imaging sensor. Physical modeling for each factor is made from which the definitive color representation will be straightforwardly derived.

First, we assume that the source illuminations are Planckian which could cover most lighting conditions such as daylight and LED lamps, i.e. the spectral radiance of lighting could be formulated by $B(\lambda,T)= \frac{2hc^{2}}{\lambda^{5}}\frac{1}{e^{hc/\lambda k_BT}-1}$ where $h=6.626 \times 10^{-34}J\cdot s$ and $k_B=1.381\times 10^{-23}J\cdot k^{-1}$ are the Planck constant and the Boltzmann constant, respectively; $\lambda$ characterizes the lighting spectrum; $T$ represents the lighting color temperature and $c=3\times 10^8m\cdot s^{-1}$ gives the speed of light in the medium. Additionally, since the visible spectrum for the human eye always falls on high frequencies where $hc/\lambda\gg k_BT$,  the spectral power distribution $E(\lambda,T)$ of illumination with an overall intensity $I$ tends to Wien's approximation \cite{wyszecki2000color}:
\begin{equation}\label{spd}
E(\lambda,T)\simeq I\frac{k_1}{\lambda^{5}}e^{-\frac{k_2}{\lambda T}}
\end{equation}
where $k_1=2hc^2$ and $k_2=\frac{hc}{k_B}$ refer to first and second radiation constants. Moreover, as proven in \cite {finlayson2009entropy}, the Planckian characteristic can be approximately considered linear which allows us to generalize this assumption to a bi-illuminant or multi-illuminant scene.
 
The assumption for skin surface is already made, i.e. the skin is a Lambertian surface and it follows the reflection rule specified in \eqref{Lambert}. With the sensor response curve $F_i(\lambda)$ corresponding to three color channels, the spectral reflectance function of skin surface $S(\lambda)$ and aforementioned spectral power distribution $E(\lambda)$, the final output of camera sensors in RGB channels \boldmath $C$ \unboldmath $=\{R,G,B\}$ could be represented as an integral of their product over the spectrum:
\begin{equation}\label{response1}
C_i=\int F_i(\lambda) E(\lambda) S(\lambda) (\boldsymbol{n_k \cdot l}) d\lambda,\ \ \ i=1,2,3
\end{equation}
where $(\boldsymbol{n_k \cdot l})$ describes the inner product between surface normal and illumination direction. Given a specific scene and geometry, this product value for each surface point is fixed to a constant $\alpha$. 

A widely used assumption in computer graphics, which  is subsequently adopted here, is that camera sensors are sharp enough and that their spectral sensibility could be characterized by Dirac delta function $F_i(\lambda) = f_i\delta(\lambda - \lambda_i)$, which satisfies $\int F_i(\lambda) d\lambda = f_i$ and turns the integral representation in \eqref{response1} to a multiplicative form in \eqref{response2}.
\begin{equation}\label{response2}
C_i = \alpha f_i E(\lambda_i) S(\lambda_i),\ \ \ i=1,2,3
\end{equation}

Eventually, a comprehensive representation of color formation emerges after combination of \eqref{spd} and \eqref{response2}:
\begin{equation}\label{response3}
C_i = \alpha I k_1 f_i \lambda_i^{-5} e^{-\frac{k_2}{\lambda_i T}} S(\lambda_i),\ \ \ i=1,2,3
\end{equation}

An apparent truth about this formula is that the color value for one skin surface point can be practically compartmentalized into three segments: a constant part ($\alpha I k_1$), a lighting ($T$) invariant yet channel ($\lambda_i$) related part ($f_i \lambda_i^{-5} S(\lambda_i)$) and a lighting related part ($e^{-\frac{k_2}{\lambda_i T}}$). This thought-provoking observation instantly reminds us of first carrying out some normalization processing to remove the constant part and then attempting to separate the channel related part and the lighting related part for further lighting normalization. Not surprisingly, the property of intensity normalization in chromaticity space, together with the attendant investigation of the chromaticity invariant image, have come into our sight.

\section{Chromaticity Invariant Image}
The target of inferring an illumination-invariant face image based upon previously derived skin model in chromaticity space is discussed and realized in this section. We first recall the definition of chromaticity, whereafter an intrinsic characteristic of the chromaticity image in log space is studied, which leads to the following gray-scale chromaticity invariant face image formation.

\subsection{Skin Model in Chromaticity Space}

Chromaticity \cite {finlayson2009entropy, funt1992recovering, macleod1979chromaticity}, generally considered as an objective specification of the quality of color regardless of its luminance, is always defined by intensity normalized affine coordinates with respect to another tristimulus color space, such as CIEXYZ or RGB utilized in our case. The normalization mapping mainly contains two modalities: L1-normalization: \boldmath $c$ \unboldmath $=\{r,g,b\}=\{R,G,B\}/(R+G+B)$ or geometric mean normalization: \boldmath $c$ \unboldmath $=\{r,g,b\}=\{R,G,B\}/\sqrt[3]{R*G*B}$, in both normalization methods, all colors are regularized to equiluminous ones which helps to attenuate the effect of the intensity component.

For computational efficiency and further extension, the geometric-mean-normalized chromaticity is implemented as a processing pipeline for skin color in \eqref{response3}. The \boldmath $c$ \unboldmath $=\{r,g,b\}$ values in chromaticity space are given as follows:
\begin{equation}\label{chromaticity3d}
c_i = \frac{f_i \lambda_i^{-5} S(\lambda_i)}{(\prod\limits_{j=1}^{3} f_j \lambda_j^{-5} S(\lambda_j))^{\frac{1}{3}}}\frac{e^{-\frac{k_2}{\lambda_i T}}}{e^{\frac{1}{3}\sum\limits_{j=1}^{3}-\frac{k_2}{\lambda_j T}}},\ \ \ i=1,2,3
\end{equation}

Within this chromaticity representation, all constant terms are normalized. The remaining two terms consist of a channel-related one and a lighting-related one. If we switch our focus back to the process of highlight detection in the previous section which aims at separating specular reflection from diffuse reflection, the explanation could be sufficiently given: only under the assumption of the Lambertian model can we be capable of normalizing the constant terms benefiting from the multiplicative representation of skin color.

So far, we solidify and parametrize an exhaustive color formation model in a concise form. More specifically, this representation could be naturally considered as an aggregation of a lighting-invariant part and another lighting-related part, which grants us the opportunity to further explore the illumination invariant components.

\subsection{Chromaticity Invariant Image Generation}
When investigating the characteristics of the skin model in chromaticity space, both its multiplicative form and the exponential terms easily guide us to the logarithm processing, which is capable of transforming \eqref{chromaticity3d} to:
\begin{equation}\label{chromaticity3dlog}
\psi_i=\log(c_i) = \log\frac{W_i}{W}+(-\frac{k_2}{\lambda_i}-\frac{1}{3}\sum\limits_{j=1}^{3}-\frac{k_2}{\lambda_j})/T,
\end{equation}
with the lighting-invariant components $W_i=f_i \lambda_i^{-5} S(\lambda_i)$ and $W=(\prod\limits_{j=1}^{3} f_j \lambda_j^{-5} S(\lambda_j))^{\frac{1}{3}}$.

It is noticeable that all three chromaticity color channels in log space are characterized by the identical lighting color $T$ which implies the potential linear correlation among these values. Let's consider another fact: $c_1*c_2*c_3=1$ since they are geometric mean normalized values, hence it could be equally inferred that in log space we have $\ \psi_1+\psi_2+\psi_3=0$, illustrating that all chromaticity points $\boldsymbol{\psi}=(\psi_1,\psi_2,\psi_3)$ in 3D log space actually fall onto a specific plane perpendicular to its unit normal vector $\boldsymbol{u}= 1/\sqrt{3}(1,1,1)$.

Up to now, the dimensionality of target space has been reduced to 2. It becomes reasonable to bring in a 3D-2D projection in order to make the geometric significance more intuitive. Derived from the projector \boldmath $P_{u}^{\bot}=I-u^{T}u=U^{T}U$ \unboldmath onto this plane, \boldmath $U=[u_1;u_2]$ \unboldmath is a 2 $\times$ 3 orthogonal matrix formed by two nonzero eigenvectors of the projector which is able to transform the original 3D vector $\boldsymbol \psi$ to 2D coordinates $\boldsymbol \phi$ within this plane. This transformation process is portrayed in \eqref{3d2dprojection}.

\boldmath
\begin{equation}\label{3d2dprojection}
\phi = U\psi^{T}=[u_1\cdot \psi^{T};u_2\cdot \psi^{T}],
\end{equation}
\unboldmath
with $\boldsymbol{u_1}=[\frac{1}{\sqrt{2}},-\frac{1}{\sqrt{2}},0], \boldsymbol{u_2}=[\frac{1}{\sqrt{6}},\frac{1}{\sqrt{6}},-\frac{2}{\sqrt{6}}]$.

Along with the substitution of \eqref{chromaticity3dlog} in \eqref{3d2dprojection}, we are able to derive the 2D coordinates of chromaticity image pixels analytically as follows:
\begin{equation}\label{chromaticity2dlog}
\boldsymbol{\phi}=
\begin{pmatrix}
\phi_1\\ 
\phi_2\\
\end{pmatrix}=
\begin{pmatrix}
\frac{\sqrt{2}}{2}(d_1+(-\frac{k_2}{\lambda_1}+\frac{k_2}{\lambda_2})/T)\\
\frac{\sqrt{6}}{6}(d_2+(-\frac{k_2}{\lambda_1}-\frac{k_2}{\lambda_2}+\frac{2k_2}{\lambda_3})/T)\\
\end{pmatrix}
\end{equation}
with $d_1=\log(\frac{W_1}{W_2}),d_2=\log(\frac{W_1 W_2}{W_3^{2}})$.

The property of linearity in the projected plane could be straightforwardly deduced via a further analysis of \eqref{chromaticity2dlog}: 
\begin{equation}\label{linearities}
\phi_2=\frac{\sqrt{3}}{3}\frac{\lambda_1(\lambda_2-\lambda_3)+\lambda_2(\lambda_1-\lambda_3)}{(\lambda_1-\lambda_2)\lambda_3}\phi_1+d
\end{equation}
where $d$ is an offset term determined by $\{W_1,W_2,W_3\}$. Considering that $W_i$ depends merely on object surface reflectance and remains constant for a given geometry even under varying lighting conditions, the points projected onto this plane should take the form of straight lines with the same slope. Moreover, points belonging to the same material should be located on the same line and the length of each line shows the variation range of lighting with respect to this material. Accordingly, the distance between each pair of parallel lines reflects the difference between different object surface properties behind them.

\begin{figure}
\hspace{.24in}
\subfloat[]{
\label{fig:improved_subfig_a}
\centering
\includegraphics[width=0.32\linewidth]{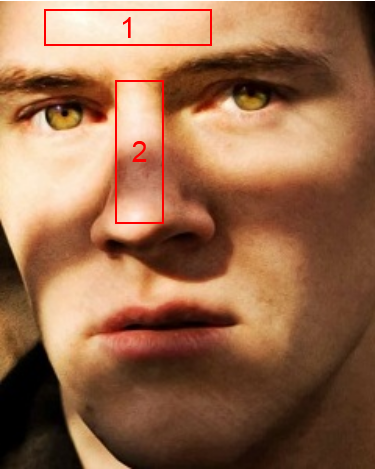}}
\hspace{.27in}
\subfloat[]{
\label{fig:improved_subfig_b}
\centering
\includegraphics[width=0.49\linewidth]{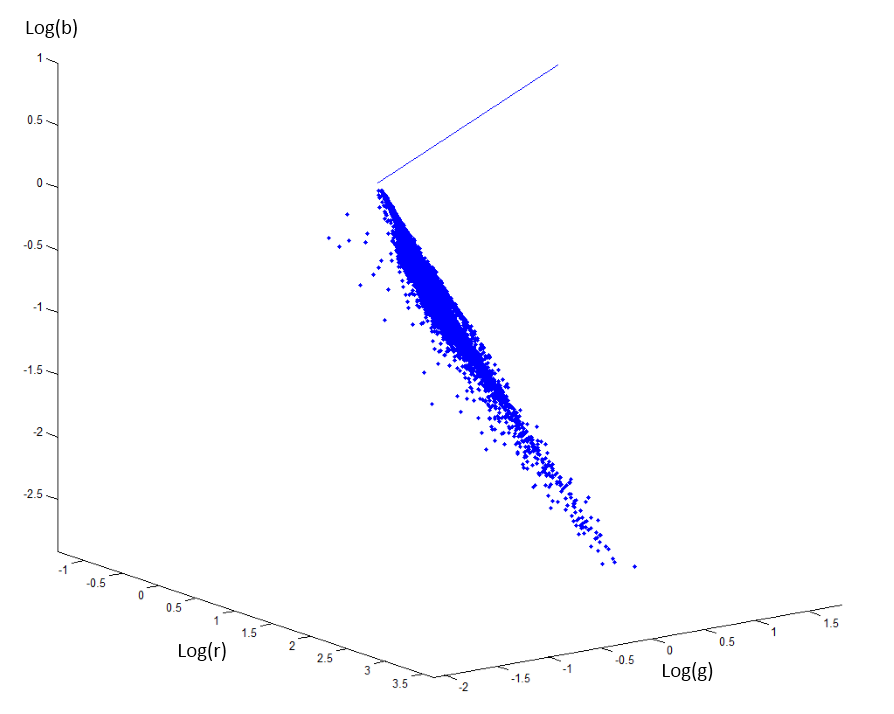}}
\\
\subfloat[]{
\label{fig:improved_subfig_c}
\centering
\includegraphics[width=0.49\linewidth]{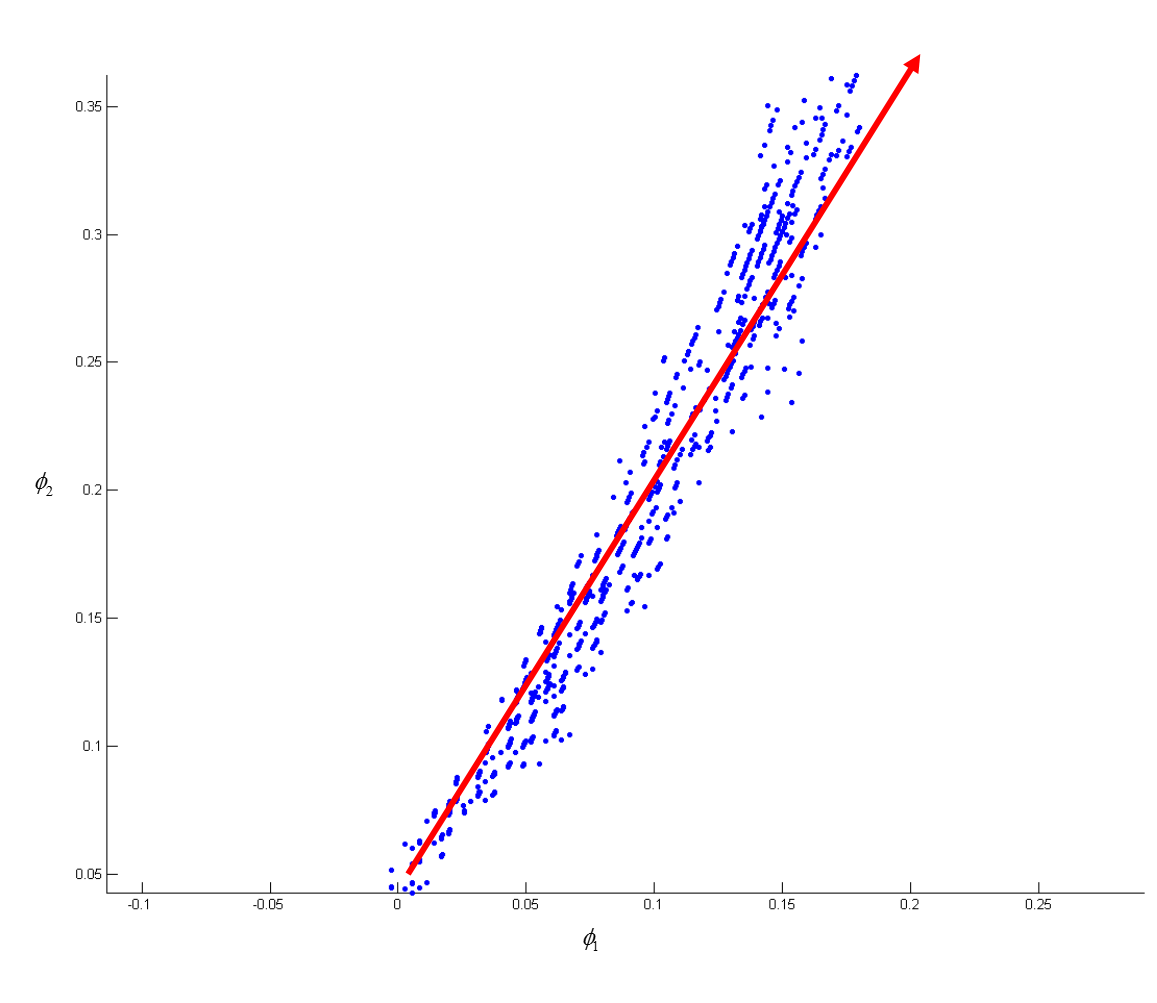}}
\subfloat[]{
\label{fig:improved_subfig_d}
\centering
\includegraphics[width=0.49\linewidth]{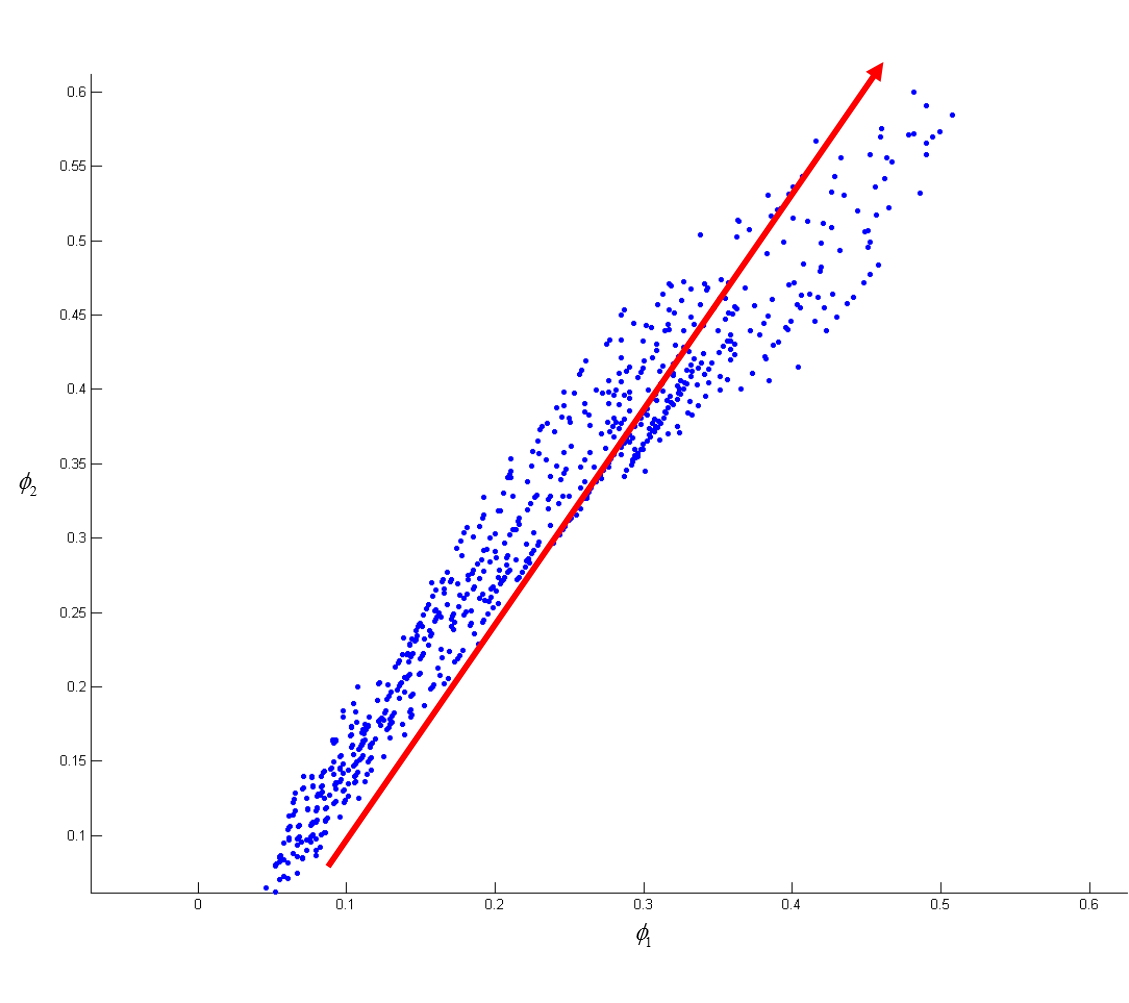}}
\caption{Linearity of chromaticity image pixels in log space. (a) Original image. (b) chromaticity pixel values in 3D log space. (c) Pixels of forehead area in projected plane. (d) Pixels of nose bridge area in projected plane.}
\label{linearity}
\end{figure}

The above inference is evidenced and supported by illustrations in Fig. \ref{linearity}. Firstly, Fig. \ref{fig:improved_subfig_b} shows that all chromaticity image points fall onto the same plane of which the normal vector, depicted with a fine blue line, is ${u}= 1/\sqrt{3}(1,1,1)$; then, we choose two sub-regions in the original image for the linearity study since the whole image contains excessive points for demonstration. Fig. \ref{fig:improved_subfig_c} and Fig. \ref{fig:improved_subfig_d} respectively represent the projected 2D chromaticity pixels in forehead and nose bridge rectangles where two approximately parallel line-shaped clusters can be obviously observed. In particular, the chosen nose bridge area bears more lighting changes while there is only unchanged directional lighting in the forehead area for comparative analysis. Correspondingly, the straight line in Fig. \ref{fig:improved_subfig_c} holds a smaller range than that in Fig. \ref{fig:improved_subfig_d}.

\subsection{Entropy based Lighting Normalization}

Note that all 2D chromaticity image pixels are scattered into line-shaped clusters differentiated by their corresponding surface attributes. To estimate the intrinsic property of different materials in chromaticity images, we would like to further reduce the dimensionality of chromaticity space.

According to \cite{barron2015shape}, global parsimony priors on reflectance could hold as a soft constraint. Under this assumption, only a small number of reflectances are expected in an object-specific image, and we reasonably extend this assumption to our own work which implies that lighting normalization substantially decreases the probability distribution of disorder in a human face image. Within this pipeline, we seek for a projection direction, parametrized by angle $\theta$, which should be exactly perpendicular to the direction of straight lines formed on the projected plane. Inasmuch as points of the same material across various illuminations fall on the same straight line, the 2D-1D projection of them onto a line with angle $\theta$ will result in an identical value which could be literally treated as an intrinsic value of this material. During this 2D-1D projection formulated in \eqref{projection2d1d}, chromaticity image is finally transformed to a 1D gray-scale image.
\begin{equation}\label{projection2d1d}
\chi = \phi_1\cos\theta+\phi_2\sin\theta
\end{equation}

With this in mind, the most appropriate projection direction could be found by minimizing the entropy of projected data. To begin with, we adopt Freedman-Diaconis rule \cite {freedman1981histogram} for the purpose of deciding the bin width as $h = 2\frac{Q(\chi)}{n^{1/3}}$, here $n$ refers to the number of projected points. Compared with the commonly used Scott's rule, Freedman-Diaconis rule replaces the standard deviation of data by its interquartile range, denoted by $Q(\chi)$, which is therefore more robust to outliers in data. Then for each candidate projection direction, the corresponding Shannon entropy can be calculated based on the probability distribution of the projected points.

\begin{figure}
\centering
\includegraphics[width=\linewidth]{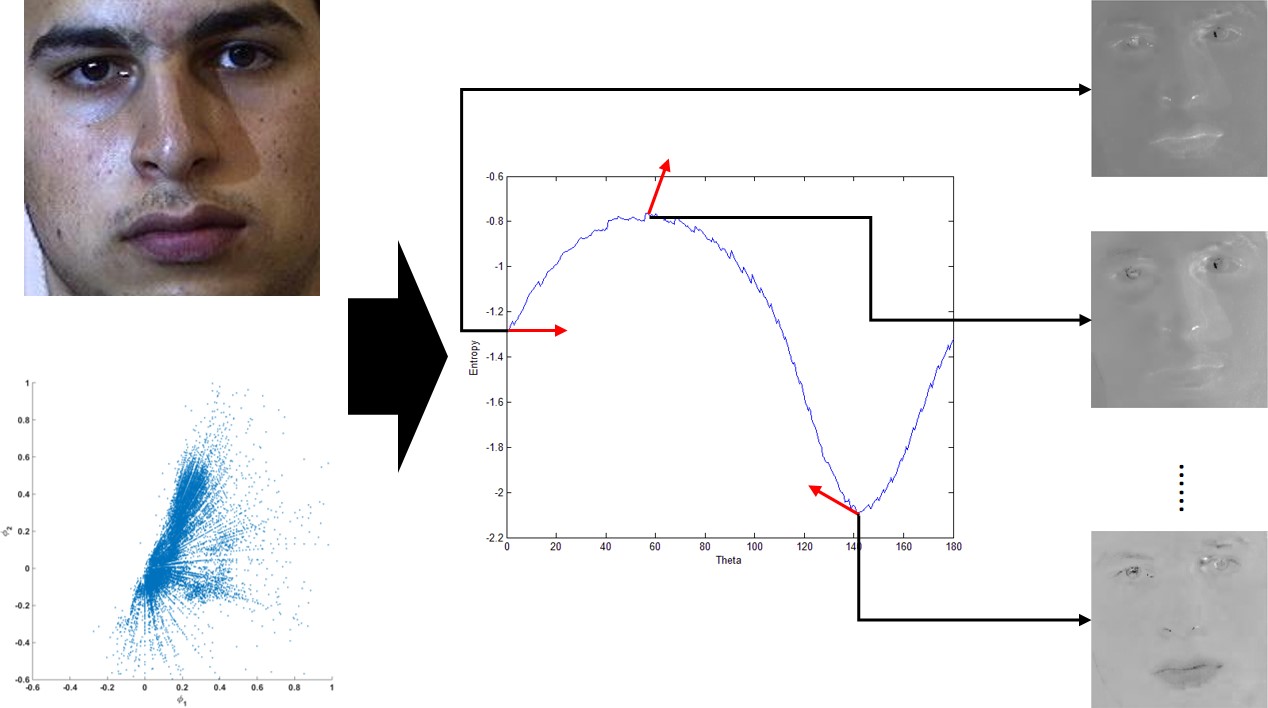}
\caption{Overview of chromaticity invariant image generation. Left column: original face image and its chromaticity points in 2D log space; middle column: entropy diagram as a function of projection angle, the arrows in red indicate projection directions at that point; right column: generated chromaticity images with different angle values.}
\label{invariant}
\end{figure}

Fig. \ref{invariant} shows the workflow of chromaticity invariant image extraction in log space. Note that we choose three different angle samples, including the zero point and two points leading to the minimum and maximum of entropy, to visualize their generated chromaticity images. Apparently, only when the angle is adjusted to the value at which the entropy comes to its minimum is shadow effect significantly suppressed in its corresponding chromaticity image, i.e. the chromaticity invariant image.  

Other than traversing all possible $\theta$ ranging from $0$ to $\pi$ inefficiently, we take an additional analysis on the slope value of projected straight lines in \eqref{linearity}, indicated by $k=\frac{\sqrt{3}}{3}\frac{\lambda_1(\lambda_2-\lambda_3)+\lambda_2(\lambda_1-\lambda_3)}{(\lambda_1-\lambda_2)\lambda_3}$. The theoretical value of slope is determined by trichromatic wavelengths $\{\lambda_1,\lambda_2,\lambda_3\}$, alternatively, the wavelengths of $\{R,G,B\}$ lights wherein $\{\lambda_1 \in [620, 750] ,\lambda_2 \in [495, 570],\lambda_3 \in [450, 495], unit: nm\}$. With simple calculations, it is interesting to find that no matter how these wavelengths change, $k$ is always a positive value and the range of $\theta$ can therefore be restricted to $[\pi/2,\pi]$ which helps to greatly reduce the computational load.

\subsection{Global Intensity Regularization}

Notwithstanding the illumination normalization, projected shadow-free images may suffer from global intensity differences across images caused by original lighting conditions and by outliers. A final global regularization module is consequently integrated in order to overcome this drawback. In this step, the most dominant intensity of the resulting image is first approximated by a simple strategy:
\begin{equation}\label{formula13}
\mu=(mean(\chi(x,y)^{m}))^{m}
\end{equation}
where $m$ is a regularization coefficient which considerably decreases the impact of large values. We take $m=0.1$ by default following the setup in the work of Tan et al. \cite {tan2010enhanced}. Next, this reference value is chosen to represent the color intensity of most face skin area and is scaled to 0.5 in a double-precision gray-scale image with data in the range [0,1]. The same scale ratio is then applied to all pixels to gain the final image.

\section{Shadow-free Color Face Recovery}
Though the representation of the 1D chromaticity invariant image contains successfully normalized lighting variations across the whole face image, it is flawed due to the loss of textural details during the process of dimensionality reduction which leads to low contrast images as depicted in Fig. \ref{invariant}. A full color image reconstruction module is therefore required to both improve the realism of generated images and enhance the performance in face analysis.

\subsection{In-depth Analysis of 1D Chromaticity Image}
Given a chromaticity invariant image and all projection matrices, a general idea to reconstruct its color version is to project reversely its 1D lighting-normalized points to 2D/3D space in steps. However, this solution is virtually impracticable due to two facts: 1) the recovery of overall intensity in each color band is an ill-posed problem since the shadow removal method is designed only for chromaticity values, 2) considerable textural features, such as the mustache and the eyebrow, are undesirably eliminated or wrongly recognized as being skin during the forward 2D/1D projection. Thus an extra analysis on representation of RGB channels in log space is conducted.

Derived from equation \eqref{response3}, the logarithmic representation of RGB values, denoted by $L_i$, could be written as a two-component addition:
\begin{equation}\label{logresponse}
L_i = \log(\alpha I k_1 f_i \lambda_i^{-5} S(\lambda_i)) - \frac{k_2}{\lambda_i T},\ \ \ i=1,2,3
\end{equation}

It is worth noting that the first additive component in the above equation consists of spatially varying factors while the second additive term is lighting-dependent. Given an illumination-invariant region, the gradients at pixel (x,y) are then computed during inference:
\begin{equation}\label{gradient}
\begin{aligned}
\nabla_x{L_i(x,y,T)} = \frac{L_i(x+\Delta{x},y,T)-L_i(x,y,T)}{\Delta{x}} \\
\nabla_y{L_i(x,y,T)} = \frac{L_i(x,y+\Delta{y},T)-L_i(x,y,T)}{\Delta{y}}
\end{aligned}
\end{equation}

Based on evidence in \cite{finlayson2006removal} and \cite{land1971lightness}, lighting conditions change slowly across a face image except for shadow edges. Consequently, for the partial derivative of the log-image with respect to $x$ at any pixel $(x,y)$ which appears out of shadow edges we have:
\begin{equation}\label{lightinginvariant}
\nabla_x{L_i(x,y,T_1)} = \nabla_x{L_i(x,y,T_2)}, \forall  (T_1,T_2)
\end{equation}
where $T_1$ and $T_2$ refer to different lighting conditions such as illuminated part and shadow part and this property holds equally for the partial derivative with respect to y.

To summarize, lighting conditions across a log-image are mainly changed on the boundary of shadow area, i.e. for any pixel inside or outside this boundary, the spatial gradient is practically lighting-invariant. Motivated by this, we will derive a shadow-specific edge detection method analytically.

\subsection{Shadow-Specific Edge Detection}
The ability to separate out shadow-specific edges from edges between different facial parts is crucial. To achieve this aim, we trace back the generation of the 1D chromaticity invariant image, where the shadow edges are removed by an orthogonal projection. Note that this projection was determined by an angle $\theta_{min}$ which minimizes the entropy of \eqref{projection2d1d}. Conversely, a 'wrong' projection angle would retain or even highlight the shadow edge.

More specifically, we seek a novel direction $\theta_{max}$ along which the projection of chromaticity pixels to 1D tends to clearly preserve the chaos caused by varying lighting conditions. The $\theta_{max}$ could be estimated by maximizing the entropy. Theoretically, the freshly projected 1D image contains edges caused by both facial features and lighting variations, thus would be considered to be different from the chromaticity invariant image in order to obtain the shadow-specific edge mask $M(x,y)$. 

Furthermore, considering that lighting effects could be specially enhanced in one of the two dimensions described in \eqref{chromaticity2dlog}, we define $M(x,y)$ as follows while combining comparisons in both re-projected $\phi_1^{min}, \phi_2^{min}$ and $\phi_1^{max}, \phi_2^{max}$:
\begin{equation}\label{shadowedgedetector}
M(x,y) = \left\{
\begin{array}{rl}
1 & if \ \Vert\phi^{'}_{min}\Vert<\tau_1 \ \& \ \Vert\phi^{'}_{max}\Vert>\tau_2 \\
0 & otherwise
\end{array} \right.
\end{equation}
where $\Vert\phi^{'}_{min}\Vert = max(\Vert\nabla\phi_1^{min}\Vert,\Vert\nabla\phi_2^{min}\Vert)$, $\Vert\phi^{'}_{max}\Vert = max(\Vert\nabla\phi_1^{max}\Vert,\Vert\nabla\phi_2^{max}\Vert)$ and $\tau_1, \tau_2$ are two pre-defined thresholds. 

It is worth mentioning that all 2D chromaticity images derived from both $\theta_{max}$ and $\theta_{min}$ are preprocessed by guided filter \cite{he2010guided} to facilitate the gradient calculation on a smoother version. As regards the choice of guided filter, we use matrix of ones for the chromaticity invariant image to average the intensity. Conversely, the chromaticity image with shadows will take itself for guided filtering to enforce the gradient map.

\subsection{Full Color Face Image Reconstruction}
Inasmuch as shadow edge mask is provided by the above detector, our focus can now be turned to the full color face image recovery. The algorithm simply continues the assumption that illumination variations mainly take place in the shadow edge area and could be ignored in other regions, i.e. the key to reconstructing an illumination-normalized color image is the reconstruction of a novel gradient map excluding the shadow-specific gradients.

To address this problem, we define a shadow-free gradient map $\zeta(x,y)$ for each log-RGB channel $i$ as follows:
\begin{equation}\label{newgradient}
\zeta_{k,i}(x,y) = \left\{
\begin{array}{ll}
\nabla_k L_i(x,y) & if \  M(x,y) = 0 \\
0 & otherwise
\end{array} \right.
\end{equation}
with $k \in \{x,y\}$. Apparently this novel shadow-free gradient map will lead us to a shadow-free Laplacian for each band:
\begin{equation}\label{laplacian}
\nu_i(x,y) = \nabla_x\zeta_{x,i}(x,y)+\nabla_y\zeta_{y,i}(x,y)
\end{equation}

This straightforwardly computed Laplacian, when combined with the shadow-free log-image $\widehat{L}$ to be reconstructed, allows us to easily define Poisson's equation:
\begin{equation}\label{poisson}
\nabla^2\widehat{L}_i(x,y) = \nu_i(x,y)
\end{equation}

Solving Poisson's equation is challenging. Two nontrivial priors are therefore imposed to make it soluble:  first, the Neumann boundary condition is adopted which specifies the derivative values on the boundary. Here we uniformly set them to zero for convenience; secondly, instead of enforcing the integrability of $\nu_i$, we simply discretize relevant terms and perform the calculation in matrix space. Importantly, given an image of size $M\times N$, the Laplacian operator $\nabla^2$, which acts essentially as a 2D convolution filter $[0,1,0;1,-4,1;0,1,0]$, is represented by a sparse matrix $\Lambda$ of size $MN \times MN$.

Let
\begin{equation}\label{discretepoisson}
D =
\begin{bmatrix}
-4 & 1 & 0 & 0 & 0 & \cdots & 0 \\
1 & -4 & 1 & 0 & 0 & \cdots & 0 \\
0 & 1 & -4 & 1 & 0 & \cdots & 0 \\
\vdots & \vdots & \vdots & \vdots & \vdots & \ddots & \vdots \\
0 & \cdots & 0 & 1 & -4 & 1 & 0 \\
0 & \cdots & 0 & 0 & 1 & -4 & 1 \\
0 & \cdots & 0 & 0 & 0 & 1 & -4
\end{bmatrix}
\end{equation}
and $I$ denotes an $M\times M$ unit matrix. We have
\begin{equation}\label{discretepoisson}
\Lambda =
\begin{bmatrix}
D & I & 0 & 0 & 0 & \cdots & 0 \\
I & D & I & 0 & 0 & \cdots & 0 \\
0 & I & D & I & 0 & \cdots & 0 \\
\vdots & \vdots & \vdots & \vdots & \vdots & \ddots & \vdots \\
0 & \cdots & 0 & I & D & I & 0 \\
0 & \cdots & 0 & 0 & I & D & I \\
0 & \cdots & 0 & 0 & 0 & I & D
\end{bmatrix}
\end{equation}

Each row of $\Lambda$ corresponds to a sparse full-size filter for one pixel, and $\widehat{L}_i$ could be accordingly solved by a left division:
\begin{equation}\label{poissonsolver}
\widehat{L}_i = \Lambda \setminus \nu_i
\end{equation}

\begin{figure}
\centering
\includegraphics[width=\linewidth]{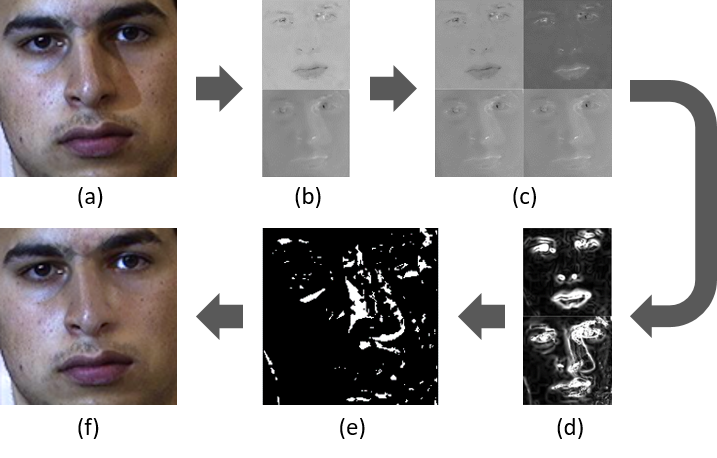}
\caption{Overview of edge mask detection and full color face recovery. (a) and (f) are raw and recovered face image; (b), (c) and (d) depict respectively 1D/2D chromaticity images and edge maps, note that in each figure the upper row refers to shadow-free version and the lower row is shadow-retained version; (e) is the final detected edge mask.}
\label{edge}
\end{figure}

After exponentiating $\widehat{L}_i$, a multiplicative scale factor per channel, which is computed by retaining the intensity of brightest pixels in raw image, will be finally applied to ensure that not only color but also intensity is properly recovered. See Fig. \ref{edge} for a demonstration of shadow-specific edge detection and color face recovery results.

\section{Experimental Results}
To quantitatively evaluate the universality and robustness of the proposed method, experiments for face recognition were carried out on several publicly available face databases, which incorporate a great deal of variety in terms of illumination environments. For each database, we adopt the standard evaluation protocols reported in the face analysis literature and present how the proposed approach improves FR performance.

\subsection{Databases and Experimental Settings} 
\textit{Databases.} In light of the fact that our method aims to normalize and recover illumination in RGB color space, two criteria need to be fulfilled in selecting a database: that it includes face images taken with various lighting conditions; and that all images are provided with full color information. The two selected databases are as follows:

\begin{itemize}
\item{The $\textbf{CMU-PIE}$ database \cite{sim2003cmu} has been very influential and prevalent in robust face recognition across pose, illumination and expression variations. It contains 41368 images from 68 identities, including 13 different poses, 43 different illumination conditions and 4 expressions. Here we restrict our attention merely to geometrically aligned frontal face views with neutral expression across illumination variations, wherein the experimental protocol in \cite {han2013comparative} is adopted.}
\item{The Face Recognition Grand Challenge ($\textbf{FRGC}$) ver2.0 database \cite{phillips2005overview} is a well-known face database designed for multitask 2D/3D FR evaluations. There are 12,776 still images and 943 3D scans from 222 identities in the training set. Accordingly, 4,007 3D scans and more than 22,000 images from 466 identities are stored in the validation set. Specifically, this database contains various scale and image resolutions as well as expression, lighting and pose variations. The standard protocol of Exp.4 defined in \cite{phillips2005overview} targeting at lighting relevant tasks is used in our FR experiments.}
\end{itemize}

\begin{figure}
\centering
\subfloat[]{
\label{fig:improved_subfig_a1}
\centering
\includegraphics[width=\linewidth]{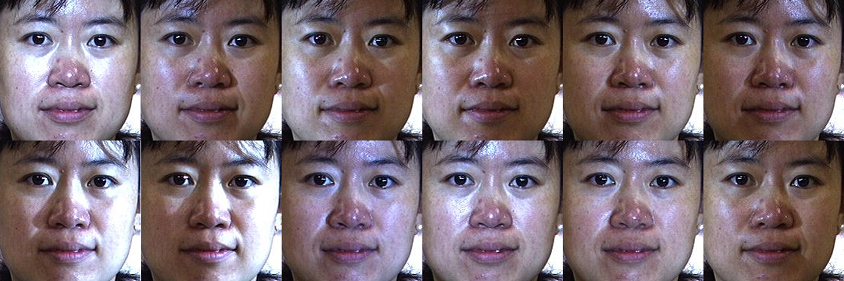}}  \\
\subfloat[]{
\label{fig:improved_subfig_b1}
\centering
\includegraphics[width=\linewidth]{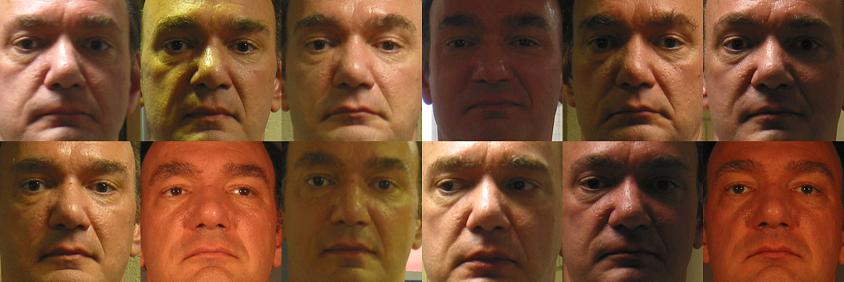}}
\caption{Cropped face examples of the first subject in the (a): CMU-PIE database; (b): FRGC database.}
\label{pie}
\end{figure}

For the first subject of each database, Fig. \ref{pie} gives an illustration of some image samples across varying illumination environments. Note that all facial images are cropped and the resolution is $180 \times 180$. As can be visualized from these figures, CMU-PIE database contains well-controlled illuminations and strictly unchanged pose for one subject while FRGC database contributes more to the variations on illumination and pose, which makes our evaluation process comprehensive and reliable.

Table \ref{table:database} gives detailed structure as well as experimental protocol for each database. According to commonly used protocols, two different tasks are proposed for these two databases: 1-v-n face identification for CMU-PIE and 1-v-1 face verification for FRGC, which will be further detailed in upcoming subsections.

\renewcommand{\arraystretch}{1.3}
\setlength{\tabcolsep}{5pt}
\begin{table}
\caption{Overview of database division in our experiments}
\label{table:database}
\centering
\begin{tabular}{| c | c | c c | c c |}
\hline
\multirow{2}{*}{Database} & \multirow{2}{*}{Person} & \multicolumn{2}{|c|}{Target Set} & \multicolumn{2}{|c|}{Query Set}\\ \cline{3-6}
& & Lighting & Images & Lighting & Images \\ \hline \hline
CMU-PIE & 68 & 3 & 204 & 18 & 1,224 \\
FRGC & 466 & controlled & 16,028 & uncontrolled & 8,014 \\
\hline
\end{tabular}
\end{table}


\textit{Features.} To evaluate performance robustness under different feature extraction algorithms, we have experimented with four popular descriptors in face recognition, including Local Binary Pattern (LBP), Local Phase Quantization (LPQ), Local Gabor Binary Pattern (LGBP) and deep CNN based face descriptor (VGG-Face), the parameter settings for each of them are detailed as follows:
\begin{itemize}
\item{LBP \cite{ahonen2006face}: For each face image a 59-dimensional uniform LBP histogram feature is extracted. For the LBP computation we set the number of sample points as 8 and radius as 2. Chi-square distance is computed between two LBP histogram features to represent their dissimilarity.}
\item{LPQ \cite{ahonen2008recognition}: We set size of the local uniform window as 5 and the correlation coefficient $\rho$ as 0.9. Accordingly, the $\alpha$ for the short-time Fourier transform equals the reciprocal of window size, i.e. $\alpha = 0.2$. With the process of decorrelation, the output feature for each image is a 256D normalized histogram of LPQ codewords and Chi-square distance is applied as well in our experiments as a matching criterion.} 
\item{LGBP \cite{zhang2005local}: For each face image, 4 wavelet scales and 6 filter orientations are considered to generate 24 Gabor kernels. Similarly to LBP, holistic LGBP features are extracted for test images, resulting in 1,416D feature vectors. A simple histogram-intersection-matching described in \cite{zhang2005local} is used as similarity measurement.}
\item{VGG-Face \cite{Parkhi15}: The VGG-Face descriptors are computed based on the VGG-Very-Deep-16 CNN architecture in \cite{Parkhi15} which achieves state-of-the-art performance on all popular FR benchmarks. Here we simply take the pre-trained model and replace the last Softmax layer by identity module in order to extract 4,096D features for test images.}
\end{itemize}

\textit{Methods.} The main contributions of our method are to remove shadows and recover illumination-normalized color face images instead of de-lighting in gray-scale like all other existing methods do. To better present the effectiveness and necessity of the proposed method, we implement it as a preprocessing followed by other gray-scale level lighting normalization techniques to test the fusion performance compared with the results obtained without using our method. As an exception to the above, for VGG-Face model which requires RGB images as input, we conduct the comparison only between original images and shadow-free recovered images with no gray-scale level lighting normalization.

For this comparative study, a bunch of gray-scale space based approaches are covered, including basic methods such as Gaussian filter based normalization (DOG), Gradient faces based normalization (GRF) \cite{zhang2009face}, wavelet based normalization (WA) \cite{du2005wavelet}, wavelet-based denoising (WD) \cite{zhang2009multiscale}, single-scale and multi-scale retinex algorithms (SSR and MSR) \cite{jobson1997properties,jobson1997multiscale}, and state-of-art methods such as logarithmic discrete cosine transform (DCT) \cite{chen2006illumination}, single-scale and multi-scale self-quotient image (SQI and MSQ) \cite{wang2004generalized}, single-scale and multi-scale Weberfaces normalization (WEB and MSW) \cite{wang2011illumination}, additionally, a well-known fusing preprocessing chain (TT) \cite{tan2010enhanced} is also experimented. Thankfully, an off-the-shelf implementation provided by \v{S}truc and Pave\v{s}i\'{c} \cite {ACKNOWL1, ACKNOWL2}, namely INface Toolbox, grants us the opportunity to achieve our target efficiently and accurately.

\begin{figure}
\centering
\includegraphics[width=0.98\linewidth]{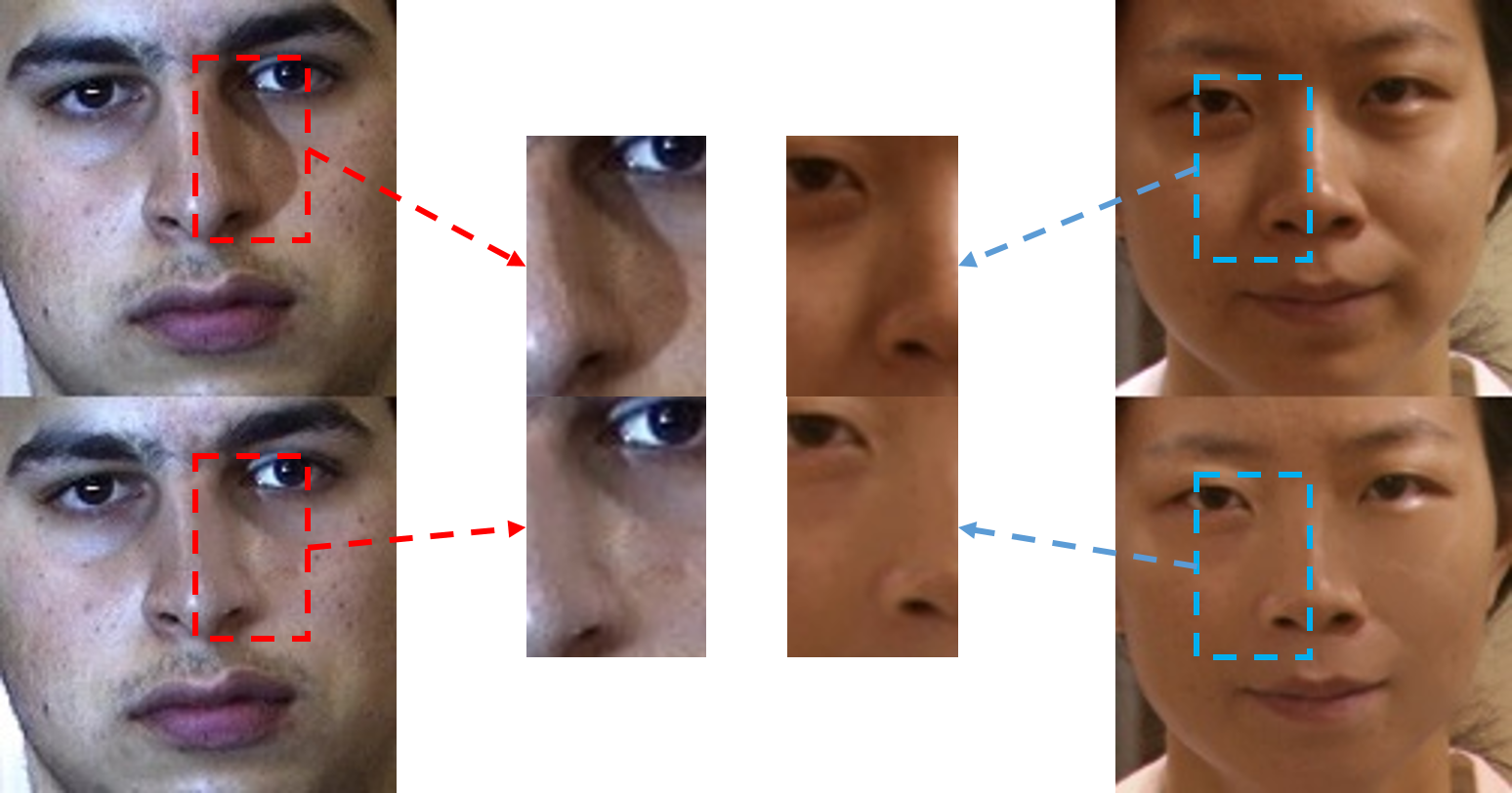}
\caption{Holistic and local shadow removal results on hard-edged shadows (left) and soft shadows (right).}
\label{compare}
\end{figure}

\begin{figure*}[t]
\centering
\subfloat[]{
\label{fig:illu_subfig_a}
\centering
\includegraphics[width=0.98\linewidth]{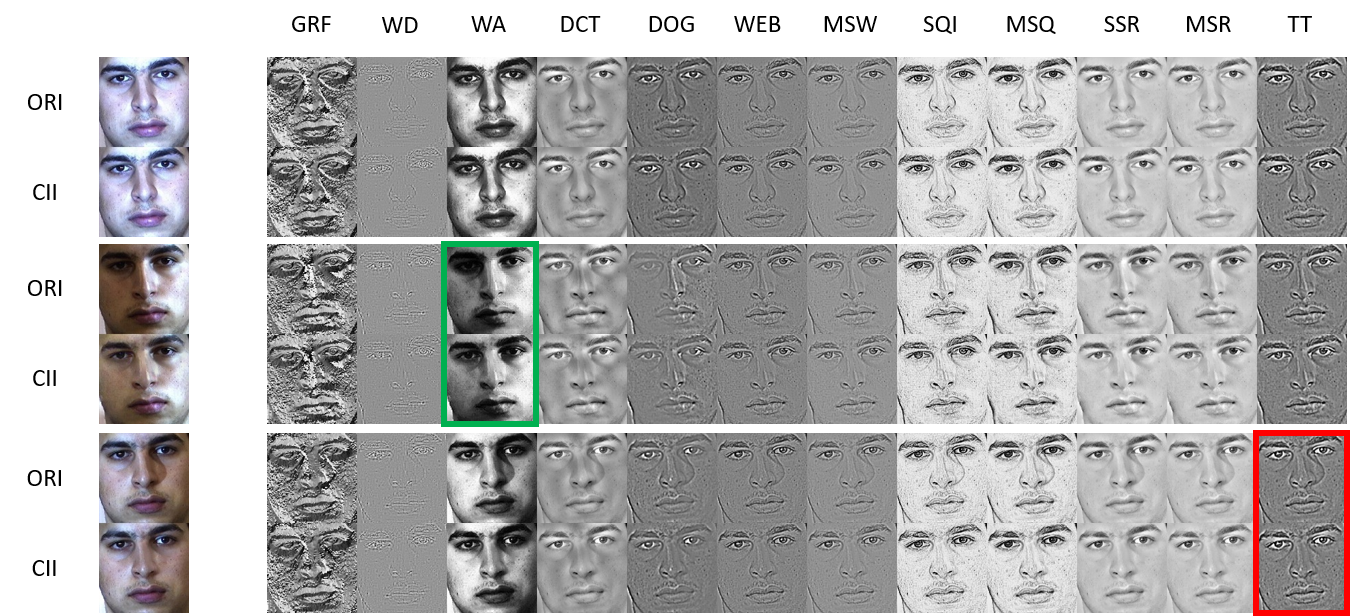}} \\
\subfloat[]{
\label{fig:illu_subfig_b}
\centering
\includegraphics[width=0.98\linewidth]{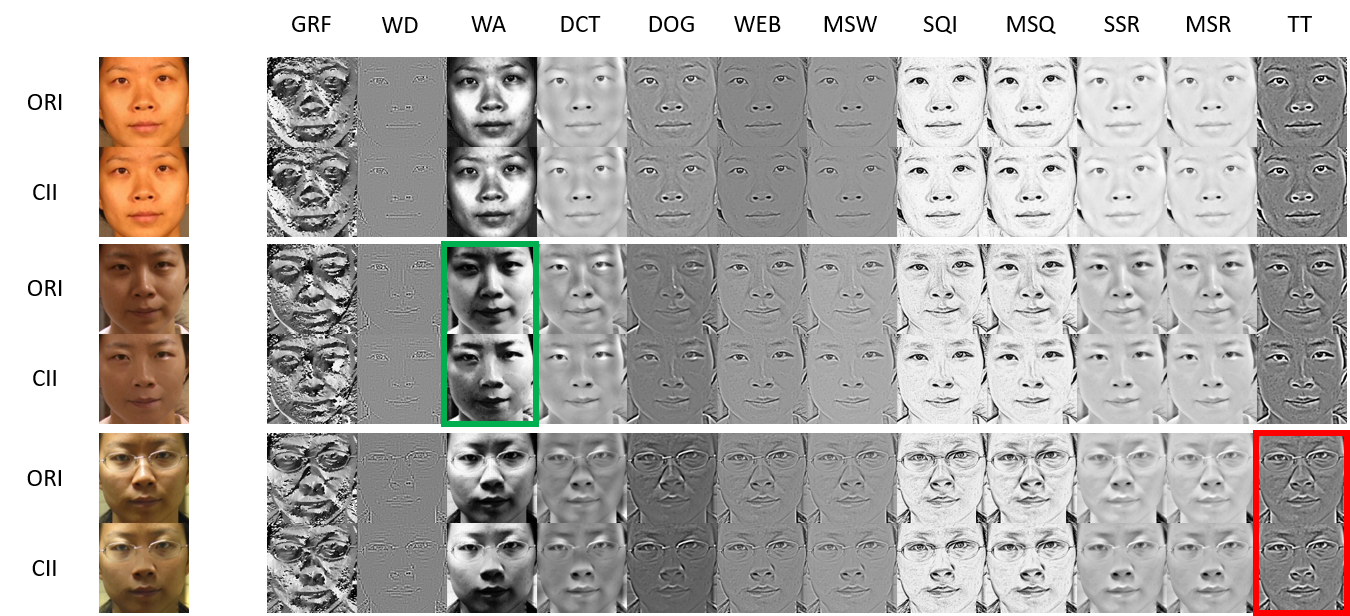}}
\caption{Illustration of illumination normalization performance of two samples in (a) CMU-PIE and (b) FRGC database. For each sample, three lighting conditions are considered, from top to bottom are the image with frontal lighting, image with soft shadows and image with hard-edged shadows.The columns represent different lighting normalization techniques to be fused with original color image or CII recovered color image.}
\label{visualization}
\end{figure*}

\subsection{Visual Comparison and Discussion}
\textit{Shadows.} Firstly, a comparison of shadow removal results on soft and hard shadows is conducted and depicted in Fig. \ref{compare}. We can make two observations from these results:
\begin{enumerate}
\item From a holistic viewpoint, our proposed method handles well the removal of both hard and soft edge shadows. In both cases, the lighting intensity across the whole image is normalized and the effects of shadows are eliminated.
\item Specified in dashed-red and dashed-blue rectangles respectively, the two middle image patches show us the differences while processing different shadows. Despite visually similar results, for face images on the left with a hard-edged shadow, shadow removal performance is actually more robust than for the image on the right with soft shadows because more facial details are smoothed for soft shadows where shadow edges are difficult to define. This drawback may also affect the performance of face recognition which will be detailed in next subsection.
\end{enumerate}

\renewcommand{\arraystretch}{1.3}
\setlength{\tabcolsep}{6pt}
\begin{table*}
\caption{Rank-1 Recognition Rates (Percent) of Different Methods on CMU-PIE Database}
\label{table:cmu_pie}
\centering
\begin{tabular}{| c | c | c c c c c c c c c c c c c|}
\hline
\multirow{2}{*}{Feature} & \multirow{2}{*}{Preprocessing} & \multicolumn{13}{c|}{Gray-Scale Lighting Normalization Methods} \\ \cline{3-15} &  & N/A & GRF & WD & WA & DCT & DOG & WEB & MSW & SQI & MSQ & SSR & MSR & TT \\ \hline \hline
\multirow{2}{*}{LBP} & Original & 44.0 & 32.8 & 20.8 & 39.2 & 72.6 & 65.4 & 59.2 & 58.9 & \textbf{61.4} & \textbf{71.8} & 66.8 & 67.7 & \textbf{67.7} \\
& CII Recovery & \textbf{48.3} & \textbf{34.1} & \textbf{23.0} & \textbf{45.6} & \textbf{75.3} & \textbf{66.0} & \textbf{59.8} & \textbf{61.0} & 61.3 & 70.9 & \textbf{68.8} & \textbf{69.8} & 65.6 \\ \hline
\multirow{2}{*}{LPQ} & Original & 58.0 & 34.7 & \textbf{37.9} & 49.6 & 85.2 & 79.1 & 73.9 & \textbf{77.6} & \textbf{74.0} & \textbf{80.2} & 84.4 & 84.8 & \textbf{82.4} \\
& CII Recovery & \textbf{62.6} & \textbf{35.1} & 35.5 & \textbf{55.3} & \textbf{86.6} & \textbf{81.0} & \textbf{75.7} & 75.1 & 73.1 & 77.1 & \textbf{88.3} & \textbf{87.4} & 82.3 \\ \hline
\multirow{2}{*}{LGBP} & Original & 75.5 & 67.8 & 84.6 & 67.2 & 97.4 & 91.3 & 99.0 & 99.4 & 99.5 & 99.2 & 98.0 & 97.8 & \textbf{97.9} \\
& CII Recovery & \textbf{77.6} & \textbf{74.6} & \textbf{84.8} & \textbf{72.1} & \textbf{98.0} & \textbf{93.6} & \textbf{99.4} & \textbf{99.7} & \textbf{99.6} & \textbf{99.4} & \textbf{99.5} & \textbf{98.4} & 96.7 \\ \hline
\multirow{2}{*}{VGG-Face} & Original & 99.7 & - & - & - & - & - & - & - & - & - & - & - & - \\
& CII Recovery & \textbf{100} & - & - & - & - & - & - & - & - & - & - & - & - \\
\hline
\end{tabular}
\end{table*}

\textit{Fusions.} To illustrate performance in an intuitive and straightforward way preceding the quantitative evaluation, consider the image samples selected from both databases and corresponding results after different lighting normalization methods in Fig. \ref{visualization}. Three gradually varying illumination scenarios are considered in our illustration, including uniformly distributed frontal lighting, a side lighting causing soft shadows and another side lighting causing some hard-edged shadows. This setting aims to evaluate the robustness of the proposed method against a wide variety of illumination environments. From the visual comparison, we see that:
\begin{enumerate}
\item In the first scenarios of both Figs. \ref{fig:illu_subfig_a} and \ref{fig:illu_subfig_b}, we hardly observe any difference between original images and recovered images. This is due to the homogeneous distribution of lighting which tends to assign zero value to most elements of the shadow-specific edge mask $M(x,y)$. In this case our algorithm makes a judgment that very few changes are required to hold this homogeneous distribution. 
\item The two middle rows in Fig. \ref{fig:illu_subfig_a} depict a face with soft shadows mainly located on the left half of it. Before applying additional lighting normalization methods, the two leftmost images show that the recovered color image successfully normalizes the holistic lighting intensity while retaining texture details. This property can also be evidenced by contrast after fusion with a diverse range of lighting normalization methods. Note that most of these techniques could handle perfectly the removal of soft shadows such as DCT, SQI, SSR and TT. For these techniques visually indistinguishable results are obtained on both original images and recovered images. On the other hand, for techniques which are less robust to soft shadows such as WA (visualized in green boxes), taking the recovered image as input enables a globally normalized lighting intensity where dark regions, especially the area around eyes, are brightened. Compared with the original image, this process gives a better visualization result. Different from the first subject in CMU-PIE, we choose a female face from FRGC with complicated illumination conditions where shadows are more scattered. Even though certain shadows still remain around mouth with our method, we can nevertheless perceive the improvement of shadow suppression on the upper half of the face.
\item The two bottom rows in Fig. \ref{fig:illu_subfig_a} and \ref{fig:illu_subfig_b} focus on hard-edged shadows caused by occlusion by the nose and glasses against the lighting direction, respectively. Under this scenario, resulting images generated by adopting the proposed recovery method as preprocessing show distinct advantages over those generated from the original image. This kind of shadow edge is difficult to remove for existing lighting normalization methods, including the state-of-art algorithm TT (visualized in red boxes), because these methods can hardly distinguish shadow edges from the intrinsic texture. 
\end{enumerate}

To summarize, according to the results of visual comparison, our shadow-free color face recovery algorithm could (1) provide intuitively identical results to original images when illumination is homogeneously distributed everywhere; (2) normalize holistic lighting in color space when soft shadows occur; (3) be performed as a supplementary measure specifically to remove hard-edged shadows before being fused with other gray-scale level lighting processing.

\subsection{Identification Results on CMU-PIE}
A rank-1 face identification task is generally described as a 1-to-n matching system, where n refers to the number of recordings in the target set, which aims to find a single identity in the target set best fitting the query face image through similarity measurement. In this scenario, closed-set identification is performed on various recognition algorithms to evaluate the robustness of our method.

Table \ref{table:cmu_pie} tabulates the identification rate for different features. For each feature and each gray-scale lighting normalization method, we compare the results before and after taking the CII recovery algorithm as preprocessing. The higher accuracy is highlighted for each comparison pair. Several observations could be made from these results:

\renewcommand{\arraystretch}{1.3}
\setlength{\tabcolsep}{6pt}
\begin{table*}
\caption{Verification Rate (Percent) at FAR = 0.1\% Using Different Methods on FRGC V2.0 Exp.4}
\label{table:frgc}
\centering
\begin{tabular}{| c | c | c c c c c c c c c c c c c|}
\hline
\multirow{2}{*}{Feature} & \multirow{2}{*}{Preprocessing} & \multicolumn{13}{c|}{Gray-Scale Lighting Normalization Methods} \\ \cline{3-15} &  & N/A & GRF & WD & WA & DCT & DOG & WEB & MSW & SQI & MSQ & SSR & MSR & TT \\ \hline \hline
\multirow{2}{*}{LBP} & Original & 1.0 & 12.8 & 3.5 & 1.1 & 6.0 & 14.5 & 18.5 & 17.7 & 18.5 & 12.3 & 3.8 & 3.9 & 15.7 \\
& CII Recovery & \textbf{1.3} & \textbf{14.8} & \textbf{5.3} & \textbf{1.5} & \textbf{6.2} & \textbf{18.8} & \textbf{23.3} & \textbf{23.1} & \textbf{25.6} & \textbf{18.0} & \textbf{5.3} & \textbf{5.9} & \textbf{20.4} \\ \hline
\multirow{2}{*}{LPQ} & Original & 1.4 & 14.2 & 7.4 & 2.0 & 6.6 & \textbf{15.3} & 18.3 & 18.8 & 13.4 & 12.0 & 6.2 & 7.5 & \textbf{21.4} \\
& CII Recovery & \textbf{2.0} & \textbf{17.6} & \textbf{7.5} & \textbf{2.5} & \textbf{6.7} & 14.9 & \textbf{19.1} & \textbf{19.7} & \textbf{16.8} & \textbf{15.2} & \textbf{7.3} & \textbf{8.1} & 20.2 \\ \hline
\multirow{2}{*}{LGBP} & Original & 13.0 & 31.0 & 18.8 & 12.7 & 28.2 & 37.0 & 37.9 & 35.7 & 29.1 & 30.9 & 27.2 & 28.2 & 38.8 \\
& CII Recovery & \textbf{16.7} & \textbf{33.2} & \textbf{25.9} & \textbf{14.3} & \textbf{29.6} & \textbf{42.4} & \textbf{38.4} & \textbf{37.0} & \textbf{31.0} & \textbf{33.1} & \textbf{29.4} & \textbf{29.9} & \textbf{44.4} \\ \hline
\multirow{2}{*}{VGG-Face} & Original & 92.5 & - & - & - & - & - & - & - & - & - & - & - & - \\
& CII Recovery & \textbf{93.6} & - & - & - & - & - & - & - & - & - & - & - & - \\
\hline
\end{tabular}
\end{table*}

\begin{figure*}
\subfloat[]{
\label{fig:roc_subfig_a}
\centering
\includegraphics[width=0.32\linewidth]{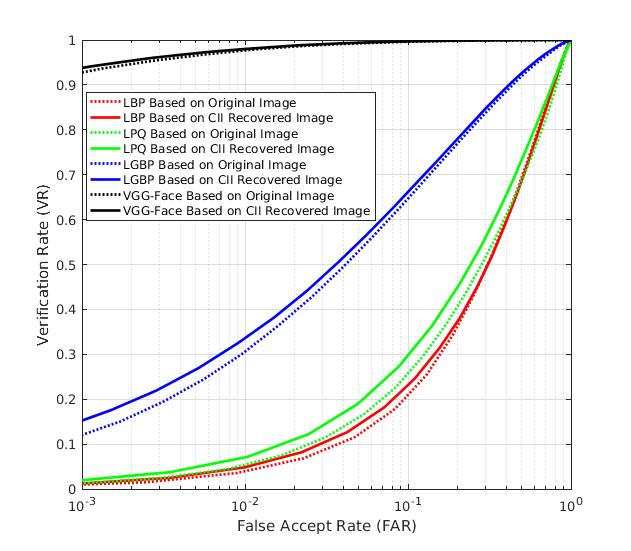}} 
\subfloat[]{
\label{fig:roc_subfig_b}
\centering
\includegraphics[width=0.31\linewidth]{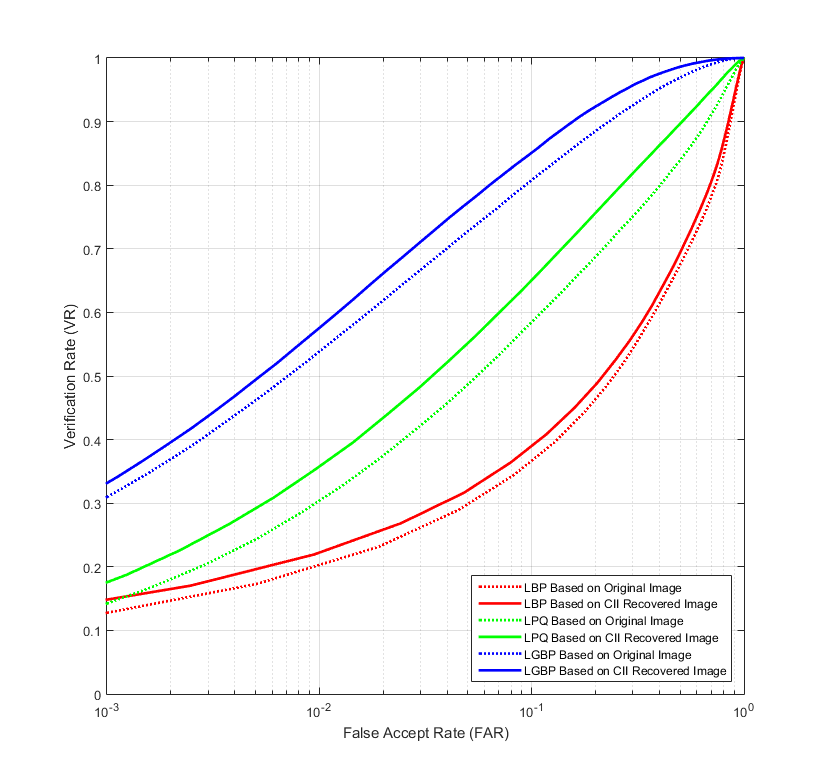}} 
\subfloat[]{
\label{fig:roc_subfig_c}
\centering
\includegraphics[width=0.305\linewidth]{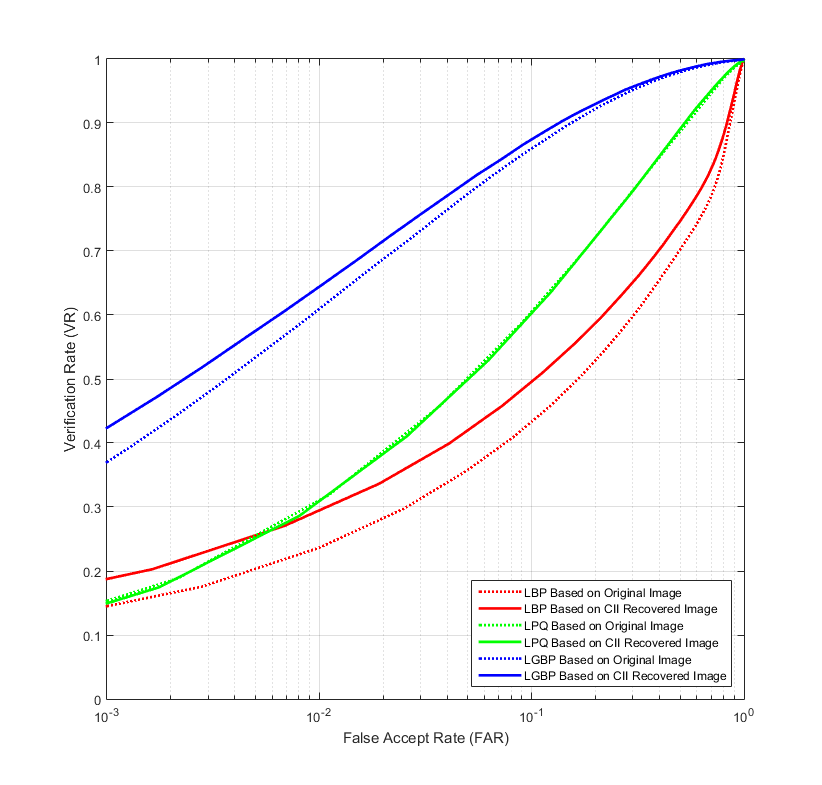}} \\
\subfloat[]{
\label{fig:roc_subfig_d}
\centering
\includegraphics[width=0.31\linewidth]{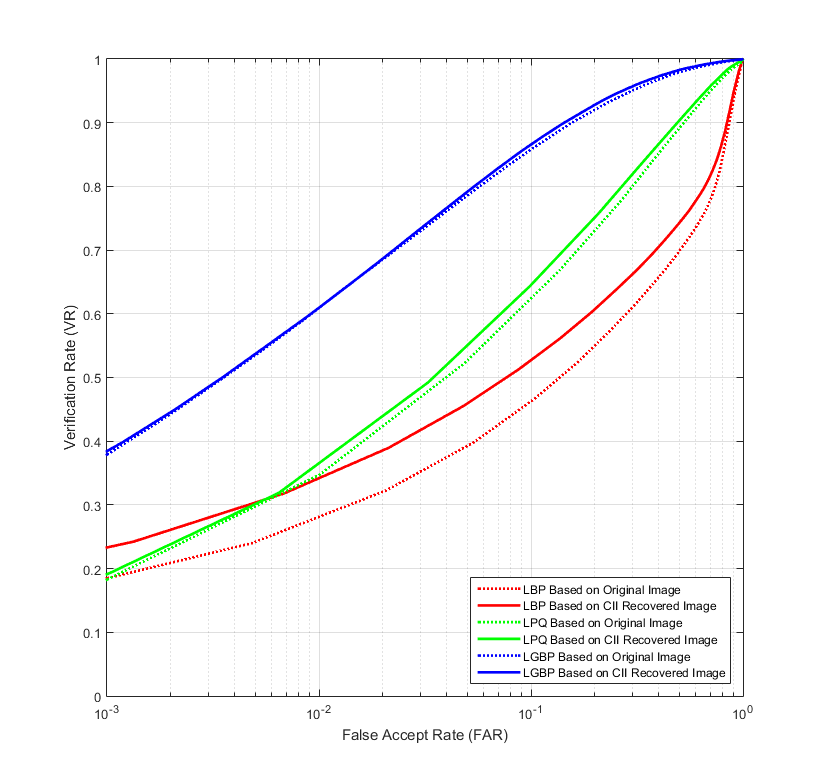}}
\subfloat[]{
\label{fig:roc_subfig_e}
\centering
\includegraphics[width=0.32\linewidth]{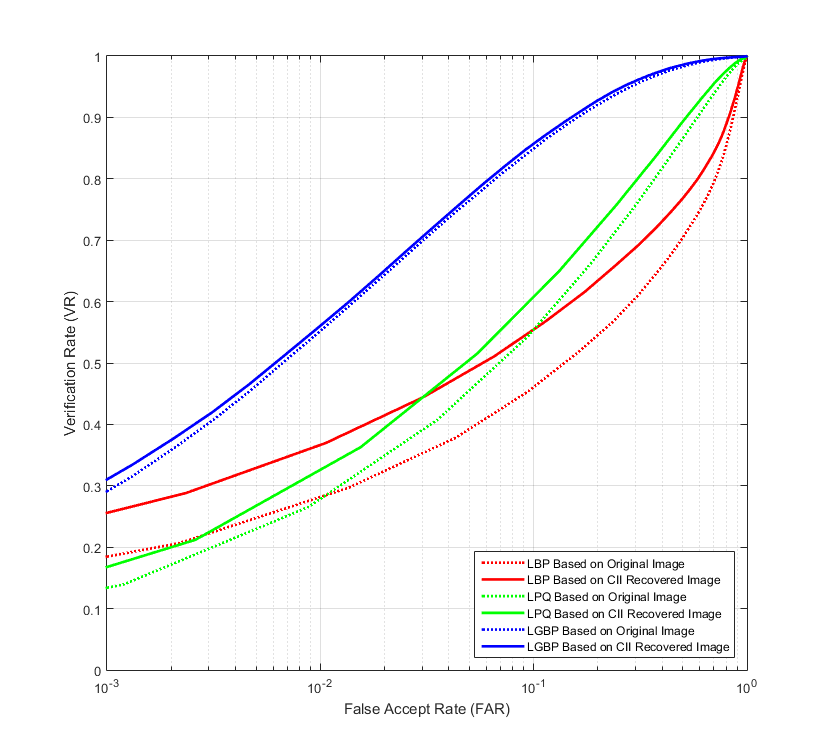}} 
\subfloat[]{
\label{fig:roc_subfig_f}
\centering
\includegraphics[width=0.32\linewidth]{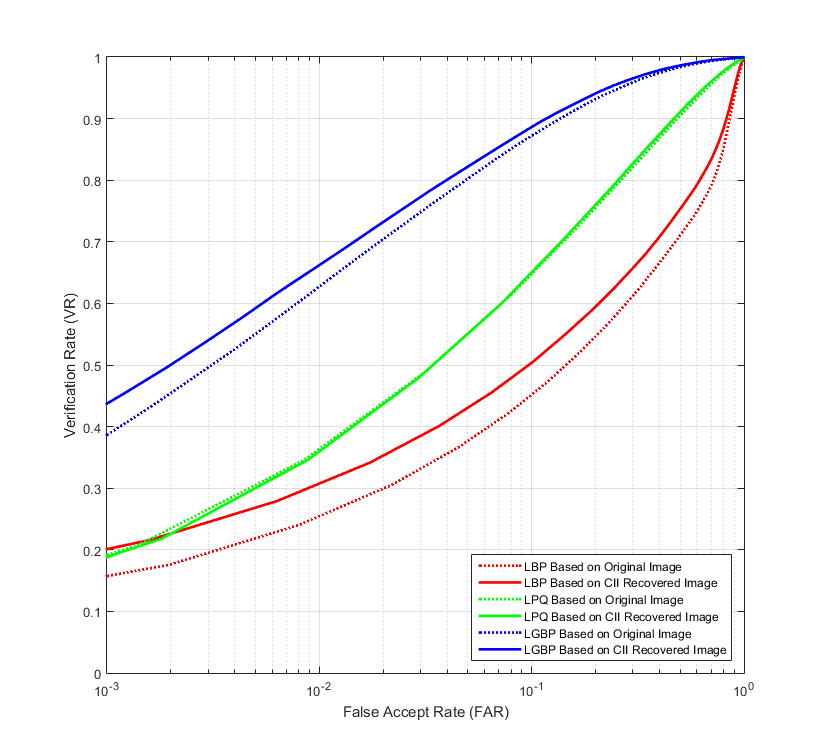}} 
\caption{Several ROC curves for different gray-scale methods. (a) No gray-scale method, (b) GRF, (c) DOG, (d) WEB, (e) SQI, (f) TT. Note that only (a) contains ROC curves for VGG-Face model because it requires RGB images as model input.}
\label{ROC}
\end{figure*}

\begin{enumerate}
\item Generally, fusing our proposed method in the preprocessing chain helps improve performance on this identification task with different gray-scale lighting normalization approaches and different features. This is because our method emphasizes the removal of shadow edges while all other methods suffer from retaining such unwanted extrinsic features.
\item Without other gray-scale methods (N/A in the Table) or even with gray-scale methods such as WA which are relatively less robust to lighting variations, the results based on the CII recovered color image significantly boost the performance compared with using other methods. This observation implies that besides the effect of shadow edge removal, our method also provides us with holistic lighting normalization as well.
\item For some gray-scale methods like SQI and MSQ, our method causes slight yet unpleasant side effects with LBP and LPQ features. This is due to the phenomenon previously observed in visual comparison that our method will smooth the detected shadow edges, SQI/MSQ may become more sensitive to this unrealistic smoothness because images would be further divided by their smoothed version. Nevertheless, with LGBP features the proposed method still achieves better results with SQI/MSQ because the introduction of 24 Gabor filters helps alleviate the effect of the smoothed region.
\item The fusion of our method and TT failed to gain performance improvement. As a preprocessing sequence itself, TT has been carefully adjusted to the utmost extent so it is difficult to combine it with other preprocessing.
\item The VGG-Face model largely outperforms the other conventional features, showing its impressive capacity in discriminative feature extraction and its robustness against holistic lighting variations. Even in this case, the implementation of the proposed method is able to further perfect the performance by dealing with the cast shadows.
\end{enumerate}

\subsection{Verification Results on FRGC}
Notwithstanding its one-to-one characteristic, face verification on FRGC v2.0 is always considered as a highly challenging task. This is due to the fact that a large number of face images in FRGC are captured in uncontrolled thus complicated illumination environments with sensor or photon noise as well. For each preprocessing combination and each feature, we conduct a 16,028 $\times$ 8,014 pair matching and then compute the verification rate based on this similarity/distance matrix. The experimental results are evaluated by Receiving Operator Characteristics (ROC), which represents the Verification Rate (VR) varying with False Acceptance Rate (FAR).

Similarly to the previous experimental setting, we list the performance of different methods on the ROC value for FAR at 0.1\% in Table \ref{table:frgc}. Moreover, corresponding ROC curves for each gray-scale method are illustrated in Fig \ref{ROC}. We make our observations from these results:
\begin{enumerate}
\item Using the recovered color image is generally an effective way to improve the performance on this verification task with different gray-scale methods and features. Compared with the identification task on CMU-PIE, this effectiveness is enhanced here since our method helps improve the verification rate at FAR = 0.1\% for almost all gray-scale methods with different features, validating the superiority of the proposed method.
\item A similar fact as in CMU-PIE is encountered again: the VGG-Face model can greatly increase performance when compared with the other features, while adding the proposed shadow removal preprocessing leads to a relatively slight (1.1\%) yet important improvement on this deep CNN model.
\item The performance variance for different gray-scale methods is not totally consistent with our previous observation on the CMU-PIE database. Unlike before, GRF, DOG and WEB achieve better results than DCT and SSR, which implies that these methods are more robust while dealing with uncontrolled and arbitrary lighting conditions.
\end{enumerate}

\section{Conclusion}
In this paper, we have presented a novel pipeline in chromaticity space for improving the performance on illumination-normalized face analysis. Our main contributions consist of: (1) introducing the concept of chromaticity space in face recognition as a remedy to illumination variations, (2) achieving an intrinsic face image extraction processing and (3) realizing a photo-realistic full color face reconstruction after shadow removal. Overall, the proposed approach explores physical interpretations for skin color formation and is proven to be effective by improving performance for FR across illumination variations on different databases. Meanwhile, it shows a promising potential in practical applications for its photo-realism and extensibility. Further efforts in developing this work will include synthesizing face images under different illumination conditions and combining pose invariant techniques in order to address face analysis problems in the wild.

\ifCLASSOPTIONcaptionsoff
  \newpage
\fi



\bibliographystyle{IEEEtran}
\bibliography{egbib.bib}

\begin{thebibliography}{10}
\providecommand{\url}[1]{#1}
\csname url@samestyle\endcsname
\providecommand{\newblock}{\relax}
\providecommand{\bibinfo}[2]{#2}
\providecommand{\BIBentrySTDinterwordspacing}{\spaceskip=0pt\relax}
\providecommand{\BIBentryALTinterwordstretchfactor}{4}
\providecommand{\BIBentryALTinterwordspacing}{\spaceskip=\fontdimen2\font plus
\BIBentryALTinterwordstretchfactor\fontdimen3\font minus
  \fontdimen4\font\relax}
\providecommand{\BIBforeignlanguage}[2]{{%
\expandafter\ifx\csname l@#1\endcsname\relax
\typeout{** WARNING: IEEEtran.bst: No hyphenation pattern has been}%
\typeout{** loaded for the language `#1'. Using the pattern for}%
\typeout{** the default language instead.}%
\else
\language=\csname l@#1\endcsname
\fi
#2}}
\providecommand{\BIBdecl}{\relax}
\BIBdecl

\bibitem{zhao2003face}
W.~Zhao, R.~Chellappa, P.~J. Phillips, and A.~Rosenfeld, ``Face recognition: A
  literature survey,'' \emph{ACM computing surveys (CSUR)}, vol.~35, no.~4, pp.
  399--458, 2003.

\bibitem{adini1997face}
Y.~Adini, Y.~Moses, and S.~Ullman, ``Face recognition: The problem of
  compensating for changes in illumination direction,'' \emph{Pattern Analysis
  and Machine Intelligence, IEEE Transactions on}, vol.~19, no.~7, pp.
  721--732, 1997.

\bibitem{pizer1987adaptive}
S.~M. Pizer, E.~P. Amburn, J.~D. Austin, R.~Cromartie, A.~Geselowitz, T.~Greer,
  B.~ter Haar~Romeny, J.~B. Zimmerman, and K.~Zuiderveld, ``Adaptive histogram
  equalization and its variations,'' \emph{Computer vision, graphics, and image
  processing}, vol.~39, no.~3, pp. 355--368, 1987.

\bibitem{shan2003illumination}
S.~Shan, W.~Gao, B.~Cao, and D.~Zhao, ``Illumination normalization for robust
  face recognition against varying lighting conditions,'' in \emph{Analysis and
  Modeling of Faces and Gestures, IEEE International Workshop on}.\hskip 1em
  plus 0.5em minus 0.4em\relax IEEE, 2003, pp. 157--164.

\bibitem{zhang2009multiscale}
T.~Zhang, B.~Fang, Y.~Yuan, Y.~Yan~Tang, Z.~Shang, D.~Li, and F.~Lang,
  ``Multiscale facial structure representation for face recognition under
  varying illumination,'' \emph{Pattern Recognition}, vol.~42, no.~2, pp.
  251--258, 2009.

\bibitem{chen2006illumination}
W.~Chen, M.~J. Er, and S.~Wu, ``Illumination compensation and normalization for
  robust face recognition using discrete cosine transform in logarithm
  domain,'' \emph{Systems, Man, and Cybernetics, Part B: Cybernetics, IEEE
  Transactions on}, vol.~36, no.~2, pp. 458--466, 2006.

\bibitem{land1971lightness}
E.~H. Land and J.~J. McCann, ``Lightness and retinex theory,'' \emph{JOSA},
  vol.~61, no.~1, pp. 1--11, 1971.

\bibitem{riklin1999quotient}
T.~Riklin-Raviv and A.~Shashua, ``The quotient image: class based recognition
  and synthesis under varying illumination conditions,'' in \emph{Computer
  Vision and Pattern Recognition, IEEE Computer Society Conference on.},
  vol.~2.\hskip 1em plus 0.5em minus 0.4em\relax IEEE, 1999.

\bibitem{wang2004generalized}
H.~Wang, S.~Z. Li, and Y.~Wang, ``Generalized quotient image,'' in
  \emph{Computer Vision and Pattern Recognition, IEEE Computer Society
  Conference on}, vol.~2.\hskip 1em plus 0.5em minus 0.4em\relax IEEE, 2004,
  pp. II--498.

\bibitem{chen2006total}
T.~Chen, W.~Yin, X.~S. Zhou, D.~Comaniciu, and T.~S. Huang, ``Total variation
  models for variable lighting face recognition,'' \emph{Pattern Analysis and
  Machine Intelligence, IEEE Transactions on}, vol.~28, no.~9, pp. 1519--1524,
  2006.

\bibitem{xie2006efficient}
X.~Xie and K.-M. Lam, ``An efficient illumination normalization method for face
  recognition,'' \emph{Pattern Recognition Letters}, vol.~27, no.~6, pp.
  609--617, 2006.

\bibitem{zhang2009face}
T.~Zhang, Y.~Y. Tang, B.~Fang, Z.~Shang, and X.~Liu, ``Face recognition under
  varying illumination using gradientfaces,'' \emph{Image Processing, IEEE
  Transactions on}, vol.~18, no.~11, pp. 2599--2606, 2009.

\bibitem{wang2011illumination}
B.~Wang, W.~Li, W.~Yang, and Q.~Liao, ``Illumination normalization based on
  weber's law with application to face recognition,'' \emph{Signal Processing
  Letters, IEEE}, vol.~18, no.~8, pp. 462--465, 2011.

\bibitem{tan2010enhanced}
X.~Tan and B.~Triggs, ``Enhanced local texture feature sets for face
  recognition under difficult lighting conditions,'' \emph{Image Processing,
  IEEE Transactions on}, vol.~19, no.~6, pp. 1635--1650, 2010.

\bibitem{basri2003lambertian}
R.~Basri and D.~W. Jacobs, ``Lambertian reflectance and linear subspaces,''
  \emph{Pattern Analysis and Machine Intelligence, IEEE Transactions on},
  vol.~25, no.~2, pp. 218--233, 2003.

\bibitem{blanz2003face}
V.~Blanz and T.~Vetter, ``Face recognition based on fitting a 3d morphable
  model,'' \emph{Pattern Analysis and Machine Intelligence, IEEE Transactions
  on}, vol.~25, no.~9, pp. 1063--1074, 2003.

\bibitem{paysan20093d}
P.~Paysan, R.~Knothe, B.~Amberg, S.~Romdhani, and T.~Vetter, ``A 3d face model
  for pose and illumination invariant face recognition,'' in \emph{Advanced
  video and signal based surveillance, 2009. AVSS'09. Sixth IEEE International
  Conference on}.\hskip 1em plus 0.5em minus 0.4em\relax IEEE, 2009, pp.
  296--301.

\bibitem{wang2009face}
Y.~Wang, L.~Zhang, Z.~Liu, G.~Hua, Z.~Wen, Z.~Zhang, and D.~Samaras, ``Face
  relighting from a single image under arbitrary unknown lighting conditions,''
  \emph{Pattern Analysis and Machine Intelligence, IEEE Transactions on},
  vol.~31, no.~11, pp. 1968--1984, 2009.

\bibitem{zhao2014minimizing}
X.~Zhao, G.~Evangelopoulos, D.~Chu, S.~Shah, and I.~Kakadiaris, ``Minimizing
  illumination differences for 3d to 2d face recognition using lighting maps.''
  \emph{IEEE transactions on cybernetics}, vol.~44, no.~5, pp. 725--736, 2014.

\bibitem{oren1994generalization}
M.~Oren and S.~K. Nayar, ``Generalization of lambert's reflectance model,'' in
  \emph{Proceedings of the 21st annual conference on Computer graphics and
  interactive techniques}.\hskip 1em plus 0.5em minus 0.4em\relax ACM, 1994,
  pp. 239--246.

\bibitem{hanrahan1993reflection}
P.~Hanrahan and W.~Krueger, ``Reflection from layered surfaces due to
  subsurface scattering,'' in \emph{Proceedings of the 20th annual conference
  on Computer graphics and interactive techniques}.\hskip 1em plus 0.5em minus
  0.4em\relax ACM, 1993, pp. 165--174.

\bibitem{phong1975illumination}
B.~T. Phong, ``Illumination for computer generated pictures,''
  \emph{Communications of the ACM}, vol.~18, no.~6, pp. 311--317, 1975.

\bibitem{belhumeur1998set}
P.~N. Belhumeur and D.~J. Kriegman, ``What is the set of images of an object
  under all possible illumination conditions?'' \emph{International Journal of
  Computer Vision}, vol.~28, no.~3, pp. 245--260, 1998.

\bibitem{ramamoorthi2001relationship}
R.~Ramamoorthi and P.~Hanrahan, ``On the relationship between radiance and
  irradiance: determining the illumination from images of a convex lambertian
  object,'' \emph{JOSA A}, vol.~18, no.~10, pp. 2448--2459, 2001.

\bibitem{wen2003face}
Z.~Wen, Z.~Liu, and T.~S. Huang, ``Face relighting with radiance environment
  maps,'' in \emph{Computer Vision and Pattern Recognition.}, vol.~2.\hskip 1em
  plus 0.5em minus 0.4em\relax IEEE, 2003, pp. II--158.

\bibitem{zhang2005face}
L.~Zhang, S.~Wang, and D.~Samaras, ``Face synthesis and recognition from a
  single image under arbitrary unknown lighting using a spherical harmonic
  basis morphable model,'' in \emph{Computer Vision and Pattern Recognition.},
  vol.~2.\hskip 1em plus 0.5em minus 0.4em\relax IEEE, 2005, pp. 209--216.

\bibitem{kee2000illumination}
S.~C. Kee, K.~M. Lee, and S.~U. Lee, ``Illumination invariant face recognition
  using photometric stereo,'' \emph{IEICE TRANSACTIONS on Information and
  Systems}, vol.~83, no.~7, pp. 1466--1474, 2000.

\bibitem{madooei2015detecting}
A.~Madooei and M.~S. Drew, ``Detecting specular highlights in dermatological
  images,'' in \emph{Image Processing (ICIP), 2015 IEEE International
  Conference on}.\hskip 1em plus 0.5em minus 0.4em\relax IEEE, 2015, pp.
  4357--4360.

\bibitem{stan2005handbook}
Z.~L. Stan and K.~J. Anil, ``Handbook of face recognition,'' \emph{Springer},
  2005.

\bibitem{hoyer2004non}
P.~O. Hoyer, ``Non-negative matrix factorization with sparseness constraints,''
  \emph{The Journal of Machine Learning Research}, vol.~5, pp. 1457--1469,
  2004.

\bibitem{wyszecki2000color}
G.~Wyszecki and W.~Stiles, \emph{Color Science: Concepts and Methods,
  Quantitative Data and Formulae}, ser. Wiley Series in Pure and Applied
  Optics.\hskip 1em plus 0.5em minus 0.4em\relax Wiley, 2000.

\bibitem{finlayson2009entropy}
G.~D. Finlayson, M.~S. Drew, and C.~Lu, ``Entropy minimization for shadow
  removal,'' \emph{International Journal of Computer Vision}, vol.~85, no.~1,
  pp. 35--57, 2009.

\bibitem{funt1992recovering}
B.~V. Funt, M.~S. Drew, and M.~Brockington, ``Recovering shading from color
  images,'' in \emph{ECCV}.\hskip 1em plus 0.5em minus 0.4em\relax Springer,
  1992, pp. 124--132.

\bibitem{macleod1979chromaticity}
D.~I. MacLeod and R.~M. Boynton, ``Chromaticity diagram showing cone excitation
  by stimuli of equal luminance,'' \emph{JOSA}, vol.~69, no.~8, pp. 1183--1186,
  1979.

\bibitem{barron2015shape}
J.~T. Barron and J.~Malik, ``Shape, illumination, and reflectance from
  shading,'' \emph{Pattern Analysis and Machine Intelligence, IEEE Transactions
  on}, vol.~37, no.~8, pp. 1670--1687, 2015.

\bibitem{freedman1981histogram}
D.~Freedman and P.~Diaconis, ``On the histogram as a density estimator: L 2
  theory,'' \emph{Probability theory and related fields}, vol.~57, no.~4, pp.
  453--476, 1981.

\bibitem{finlayson2006removal}
G.~D. Finlayson, S.~D. Hordley, C.~Lu, and M.~S. Drew, ``On the removal of
  shadows from images,'' \emph{Pattern Analysis and Machine Intelligence, IEEE
  Transactions on}, vol.~28, no.~1, pp. 59--68, 2006.

\bibitem{he2010guided}
K.~He, J.~Sun, and X.~Tang, ``Guided image filtering,'' in \emph{Computer
  Vision--ECCV 2010}.\hskip 1em plus 0.5em minus 0.4em\relax Springer, 2010,
  pp. 1--14.

\bibitem{sim2003cmu}
T.~Sim, S.~Baker, and M.~Bsat, ``The cmu pose, illumination, and expression
  database,'' \emph{Pattern Analysis and Machine Intelligence, IEEE
  Transactions on}, vol.~25, no.~12, pp. 1615--1618, 2003.

\bibitem{han2013comparative}
H.~Han, S.~Shan, X.~Chen, and W.~Gao, ``A comparative study on illumination
  preprocessing in face recognition,'' \emph{Pattern Recognition}, vol.~46,
  no.~6, pp. 1691--1699, 2013.

\bibitem{phillips2005overview}
P.~J. Phillips, P.~J. Flynn, T.~Scruggs, K.~W. Bowyer, J.~Chang, K.~Hoffman,
  J.~Marques, J.~Min, and W.~Worek, ``Overview of the face recognition grand
  challenge,'' in \emph{Computer vision and pattern recognition, 2005. CVPR
  2005. IEEE computer society conference on}, vol.~1.\hskip 1em plus 0.5em
  minus 0.4em\relax IEEE, 2005, pp. 947--954.

\bibitem{ahonen2006face}
T.~Ahonen, A.~Hadid, and M.~Pietikainen, ``Face description with local binary
  patterns: Application to face recognition,'' \emph{Pattern Analysis and
  Machine Intelligence, IEEE Transactions on}, vol.~28, no.~12, pp. 2037--2041,
  2006.

\bibitem{ahonen2008recognition}
T.~Ahonen, E.~Rahtu, V.~Ojansivu, and J.~Heikkila, ``Recognition of blurred
  faces using local phase quantization,'' in \emph{Pattern Recognition, 2008.
  ICPR 2008. 19th International Conference on}.\hskip 1em plus 0.5em minus
  0.4em\relax IEEE, 2008, pp. 1--4.

\bibitem{zhang2005local}
W.~Zhang, S.~Shan, W.~Gao, X.~Chen, and H.~Zhang, ``Local gabor binary pattern
  histogram sequence (lgbphs): A novel non-statistical model for face
  representation and recognition,'' in \emph{Computer Vision, 2005. ICCV 2005.
  Tenth IEEE International Conference on}, vol.~1.\hskip 1em plus 0.5em minus
  0.4em\relax IEEE, 2005, pp. 786--791.

\bibitem{Parkhi15}
O.~M. Parkhi, A.~Vedaldi, and A.~Zisserman, ``Deep face recognition,'' in
  \emph{British Machine Vision Conference}, 2015.

\bibitem{du2005wavelet}
S.~Du and R.~Ward, ``Wavelet-based illumination normalization for face
  recognition,'' in \emph{Image Processing, 2005. ICIP 2005. IEEE International
  Conference on}, vol.~2.\hskip 1em plus 0.5em minus 0.4em\relax IEEE, 2005,
  pp. II--954.

\bibitem{jobson1997properties}
D.~J. Jobson, Z.-U. Rahman, and G.~A. Woodell, ``Properties and performance of
  a center/surround retinex,'' \emph{Image Processing, IEEE Transactions on},
  vol.~6, no.~3, pp. 451--462, 1997.

\bibitem{jobson1997multiscale}
D.~J. Jobson, Z.-u. Rahman, and G.~A. Woodell, ``A multiscale retinex for
  bridging the gap between color images and the human observation of scenes,''
  \emph{Image Processing, IEEE Transactions on}, vol.~6, no.~7, pp. 965--976,
  1997.

\bibitem{ACKNOWL1}
V.~{\v{S}}truc and N.~Pave{\v{s}}ic, ``Photometric normalization techniques for
  illumination invariance,'' \emph{Advances in Face Image Analysis: Techniques
  and Technologies, IGI Global}, pp. 279--300, 2011.

\bibitem{ACKNOWL2}
V.~\v{S}truc and N.~Pave\v{s}i\'{c}, ``Gabor-based kernel partial-least-squares
  discrimination features for face recognition,'' \emph{Informatica (Vilnius)},
  vol.~20, no.~1, pp. 115--138, 2009.

\end{thebibliography}
\end{document}